\documentclass[11pt]{article}

\usepackage[final]{acl}

\usepackage{times}
\usepackage{latexsym}

\usepackage[T1]{fontenc}

\usepackage[utf8]{inputenc}

\usepackage{microtype}

\usepackage{inconsolata}

\usepackage{graphicx}
\usepackage{xcolor}
\usepackage{amsfonts}
\usepackage{amsmath} 
\usepackage{natbib} 
\usepackage{multirow}
\usepackage{booktabs}
\title{GUITrans2Act: Understanding User Operational Behaviors from Mobile GUI Interactions with Vision-Language Models}

\newcommand{\blfootnote}[1]{%
  \begingroup
  \renewcommand\thefootnote{}\footnote{#1}%
  \addtocounter{footnote}{-1}%
  \endgroup
}

\author{
  Yudong Zhang$^{*,1}$, Lei Hu$^{*,1}$, Daoyang Liu$^{2}$, Jiawei Liu$^{1}$, Yangfan Luo$^{1}$ \\[2pt]
  \textbf{Zhilin Gao}$^{\dagger,1}$,\textbf{Zuojian Wang}$^{\dagger,1}$ \\[8pt]
  $^{1}$\textit{Honor Device Co., Ltd} \\
  $^{2}$\textit{The Chinese University of Hong Kong, Hong Kong, China}
}

\begin{document}

\maketitle

\blfootnote{$^{*}$These authors contributed equally to this work.}
\blfootnote{$^{\dagger}$Corresponding authors.}

\begin{abstract}
  Understanding the digital world on mobile devices is shifting from static UI perception to dynamic action comprehension. This capability enables models to convert visual state transitions into operational knowledge, defined as short natural-language sentences that describe action types, target UI elements, textual arguments, and execution orders. However, due to the highly diverse and heterogeneous UI designs across applications, existing vision-language models (VLMs) struggle to accurately infer these underlying operations. To bridge this gap, we introduce Teach VLM, a core model designed to translate mobile screen trajectories into step-wise operational knowledge by extracting and analyzing operation-related keyframes from demonstration videos. To address the scarcity of aligned training data, we develop a systematic data flywheel for scalable data acquisition. We further introduce a novel Chinese Mobile Screen Teach Benchmark for fine-grained evaluation. Building upon Teach VLM, we propose the Teach-and-Repeat paradigm, where the generated operational knowledge serves as an interpretable procedural reference to guide downstream screen-based execution agents. Extensive evaluations demonstrate that Teach VLM significantly outperforms strong VLM baselines, achieving state-of-the-art performance in operation semantics prediction. Furthermore, experiments in Android World show that our paradigm yields consistent Task Success Rate improvements for downstream agents. Together, Teach VLM and the Teach-and-Repeat paradigm offer a practical pathway from raw demonstrations to reusable task automation.

\end{abstract}

\section{Introduction}

Recent multimodal large language models have made significant progress in digital-world perception, including mobile screen understanding and autonomous GUI agents~\cite{qwen3vltechnicalreport,cheng2024seeclick,wu2024atlas,rawles2024androidworlddynamicbenchmarkingenvironment,lu2025guiodyssey}. However, practical screen understanding requires more than recognizing static screenshots; a model must also explain dynamic screen changes caused by human operations and translate them into operational knowledge (i.e., short natural-language sentences describing action types, target UI elements, textual arguments, and execution orders).

\begin{figure*}[t]
    \centering
    \includegraphics[width=0.95\textwidth]{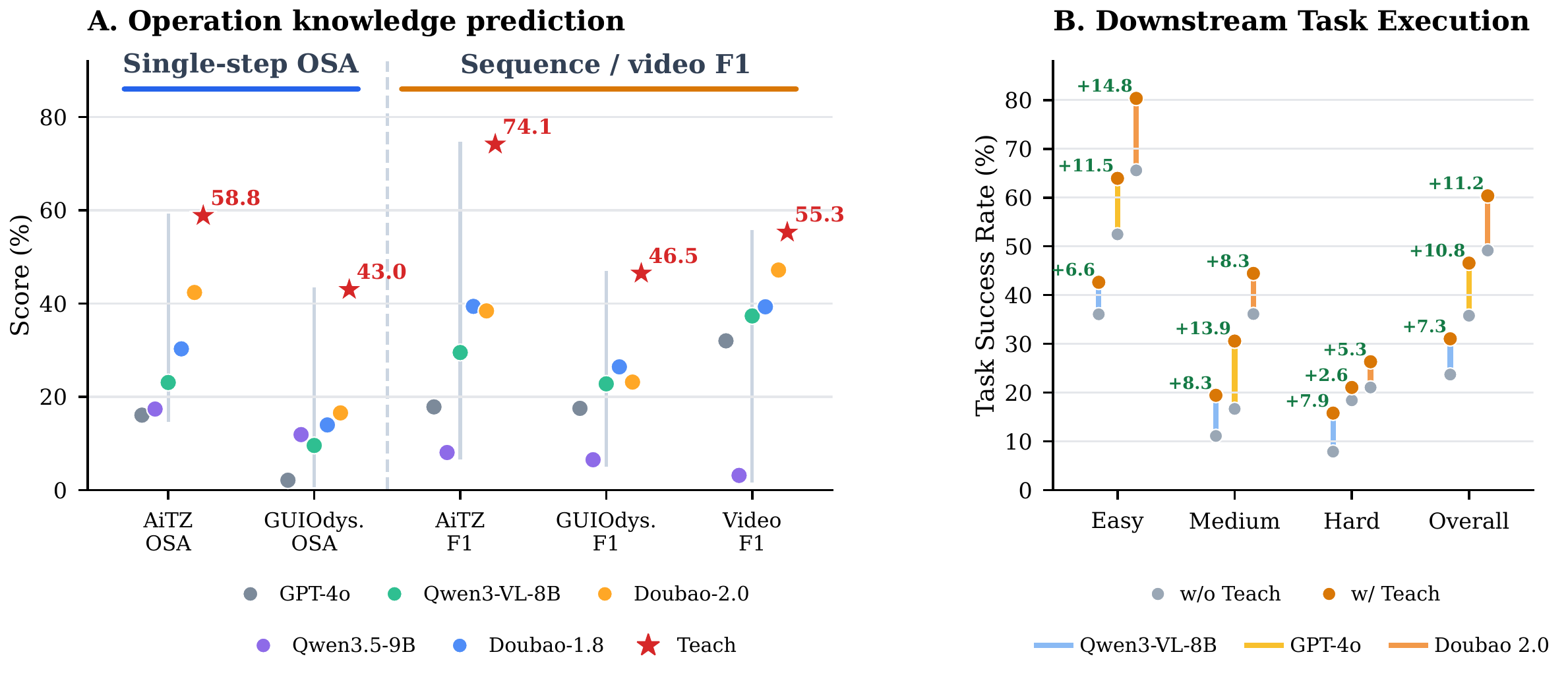}
    \caption{
    Comparison of operational knowledge extraction performance and downstream task execution effectiveness.
    Left: Teach VLM outperforms representative VLM baselines on operational knowledge prediction, including single-step operation semantic accuracy and multi-step sequence/video F1.
    Right: Teach-generated operational knowledge improves Android World Task Success Rate when injected as an external procedural reference across execution models and task difficulties.
    }
    \label{fig:quantitative_highlights}
\end{figure*}

Existing methods approach this problem from several related but incomplete directions. Static screen-understanding and grounding models focus on locating or describing visible elements~\cite{cheng2024seeclick,wu2024atlas}, but they do not directly explain the operation that links two screen states. GUI agents learn to predict the next executable action given the current screen state and a task instruction~\cite{zhang2023appagentmultimodalagentssmartphone,qin2025uitarspioneeringautomatedgui,rawles2024androidworlddynamicbenchmarkingenvironment}; however, their outputs are tied to a particular runtime state and lack interpretability. Meanwhile, modern vision-language models (VLMs) have demonstrated strong multi-frame understanding capabilities~\cite{openai2024gpt4technicalreport,qwen3vltechnicalreport}, yet they still perform poorly at recognizing operation semantics from mobile screen transitions (as shown in Figure~\ref{fig:quantitative_highlights}). This gap stems from two challenges: demonstration videos contain dense frames interspersed with task-irrelevant visual noise (e.g., loading animations and transition effects), and the interaction designs of mobile applications are highly heterogeneous across apps and platforms, preventing VLMs from generalizing operation recognition across screen contexts.

To bridge this gap, we propose Teach VLM, a core model that directly maps mobile screen transitions into operational knowledge. Given a user demonstration video, we first filter out irrelevant transitional frames to extract operation-related keyframes. Then, Teach VLM infers natural-language step-wise operation descriptions that are independent of specific apps or tasks, from the visual state changes before and after the action. Since high-quality aligned training data is extremely scarce, we design a Data Flywheel mechanism (as illustrated in Figure~\ref{fig:data_flywheel}) that forms a closed loop of model pre-annotation, human correction, and iterative re-training, enabling the low-cost construction of a large-scale, multi-domain training corpus. To enable fine-grained evaluation of the generated operational knowledge, we further introduce the Chinese Mobile Screen Teach Benchmark with frame-level semantic annotations.

\begin{figure*}[t]
    \centering
    \includegraphics[width=0.95\textwidth]{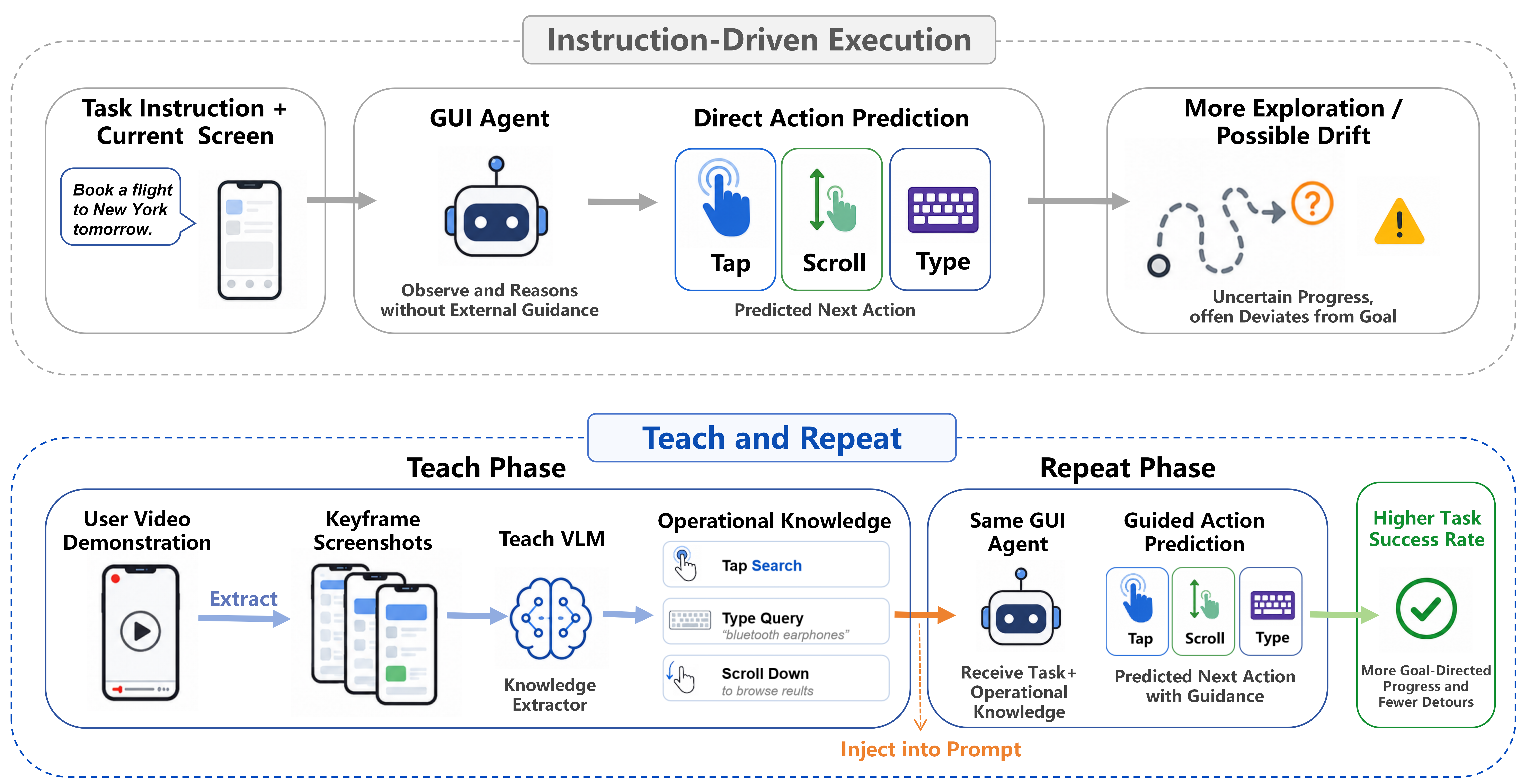}
    \caption{Overview of the proposed \textit{Teach-and-Repeat} paradigm. Compared with instruction-driven execution, which directly predicts actions from task instructions and the current screen, our framework converts a user demonstration into keyframe screenshots and uses Teach VLM to generate editable operational knowledge. The generated knowledge can be injected as an external procedural reference to guide downstream task execution.}
    \label{fig:teach_repeat_overview}
\end{figure*}

Building upon Teach VLM's capability to extract operational knowledge, we further propose the Teach-and-Repeat paradigm to address the pain points faced by downstream screen-based execution agents, such as frequent app-interface updates and ambiguous user instructions. This paradigm decouples one-time teaching from repeated execution (as illustrated in Figure~\ref{fig:teach_repeat_overview}). In the ``teach'' phase, Teach VLM converts a single user demonstration into explicit and editable operational knowledge. In the ``repeat'' phase, this natural-language knowledge is injected into execution agents as an external procedural reference. By breaking free from low-level pixel coordinates, the execution backbone can re-ground operations onto the current screen state, exhibiting strong robustness across different app versions and device states. Extensive evaluations demonstrate that Teach VLM achieves state-of-the-art performance in operation semantics prediction. Furthermore, injecting the generated knowledge as an external reference yields consistent Task Success Rate improvements for downstream execution agents in Android World (see Figure~\ref{fig:quantitative_highlights}).

In summary, our work makes the following key contributions:
\begin{itemize}
    \item \textbf{Teach VLM and Data Flywheel.} We develop a core model to accurately extract operational knowledge from mobile screen transitions, supported by a systematic data flywheel for scalable training data acquisition.
    \item \textbf{Teach-and-Repeat Paradigm.} We propose a paradigm that decouples knowledge extraction from task execution by using natural-language operational knowledge as an interpretable procedural bridge, enabling ``one-time teaching for repeated execution'' across app versions and models.
    \item \textbf{Chinese Mobile Screen Teach Benchmark.} We release a novel benchmark with frame-level semantic annotations, providing a solid foundation for fine-grained evaluation in this field.
\end{itemize}

\section{Related Work}

\paragraph{Multimodal Large Language Models.}
Robust multimodal large language model (MLLM) capabilities are essential for perceiving and interpreting screen states in the digital world~\cite{10.1093/nsr/nwae403}. Recent systems such as Gemini 1.5~\cite{reid2024gemini}, DeepSeek-V3~\cite{deepseekai2025deepseekv3technicalreport}, InternVL 3.5~\cite{wang2025internvl35advancingopensourcemultimodal}, and LLaVA~\cite{liu2023visual} demonstrate rapid progress in visual perception, instruction following, and long-context multimodal reasoning.

Among open-source MLLMs, the Qwen-VL family has become a strong foundation for vision-language applications because of its competitive visual perception, multilingual capability, and long-context modeling. Qwen3-VL~\cite{qwen3vltechnicalreport,qwen3vl8binstruct} further improves general visual understanding and provides an accessible instruction-tuned model family. Our Teach VLM is built on Qwen3-VL-8B-Instruct and adapts this general-purpose backbone to a more specific screen-transition objective. While general MLLMs can recognize visual content and answer questions about screenshots, they are not explicitly trained to recover the operation semantics that connect two mobile screen states. This work therefore focuses on converting before-and-after screen transitions into natural-language operation descriptions that can be analyzed along action, target-element, and textual-argument dimensions.

\paragraph{Screen-grounded Digital Interaction.}
Screen-grounded digital interaction studies how models perceive screen states, identify target elements, plan intermediate steps, and produce structured operations for digital tasks. This direction includes tool-use reasoning systems, mobile screen interaction models, and computer-use infrastructure. General-purpose tool-use work such as ReAct~\cite{yao2023reactsynergizingreasoningacting}, Toolformer~\cite{schick2023toolformerlanguagemodelsteach}, and GPT-4 function calling~\cite{openai2024gpt4technicalreport} provides structured interfaces for reasoning, acting, and external API invocation. Multi-agent and planning-oriented systems such as AutoGPT~\cite{yang2023autogptonlinedecisionmaking}, MetaGPT~\cite{hong2024metagptmetaprogrammingmultiagent}, and related agent surveys~\cite{masterman2024landscapeemergingaiagent} further explore online decision making and procedural decomposition. These studies show the importance of structured actions, but they usually operate over textual tools or abstract APIs rather than extracting procedural screen knowledge from visual demonstrations.

Another line of work focuses directly on screen-based interaction. AppAgent~\cite{zhang2023appagentmultimodalagentssmartphone} enables multimodal models to interact with smartphone interfaces through a simplified action space. SeeClick and OS-ATLAS~\cite{cheng2024seeclick,wu2024atlas} improve visual grounding and action modeling for screen tasks. UI-TARS~\cite{qin2025uitarspioneeringautomatedgui}, MobileAgent V3~\cite{ye2025mobileagentv3fundamentalagentsgui}, and OpenCUA~\cite{wang2025opencuaopenfoundationscomputeruse} further advance end-to-end screen interaction, long-horizon planning, and open infrastructure for computer-use research. Related datasets and environments such as Android World~\cite{rawles2024androidworlddynamicbenchmarkingenvironment} and GUIOdyssey~\cite{lu2025guiodyssey} provide dynamic tasks and cross-app navigation benchmarks. Despite this progress, most methods aim to execute actions at runtime. They must simultaneously perceive the current screen, infer the next action, and recover from unexpected states. In contrast, our work separates screen-transition understanding from execution: Teach VLM extracts operational knowledge from demonstrations, which can then be inspected, edited, evaluated, or supplied as an external procedural reference.

\paragraph{Procedural Knowledge from Interaction Experience.}
Recent studies have explored how interaction experience can be abstracted into reusable procedural guidance rather than replayed as raw trajectories. SkillRL~\cite{xia2026skillrl} evolves external procedural modules together with policy optimization, while SKILL0 and Skill-SD~\cite{lu2026skill0,wang2026skillsd} study how such guidance can serve as training-time scaffolds and be internalized into model behavior. These works suggest that interaction experience can be compressed into higher-level procedures that improve reuse and long-horizon stability.

Our work is related to this idea but differs in both granularity and input modality. Instead of learning a reusable module for a family of tasks, Teach VLM extracts step-level operational knowledge from a single mobile screen demonstration. The input is a sequence of screen-state changes, and the output is a natural-language description grounded in concrete before-and-after screenshots. This makes the knowledge more interpretable and editable, and it provides direct supervision for evaluating whether a model understands the action type, target screen element, and textual argument behind each transition.

\section{Method}

\begin{figure*}[t]
    \centering
    \includegraphics[width=1\linewidth]{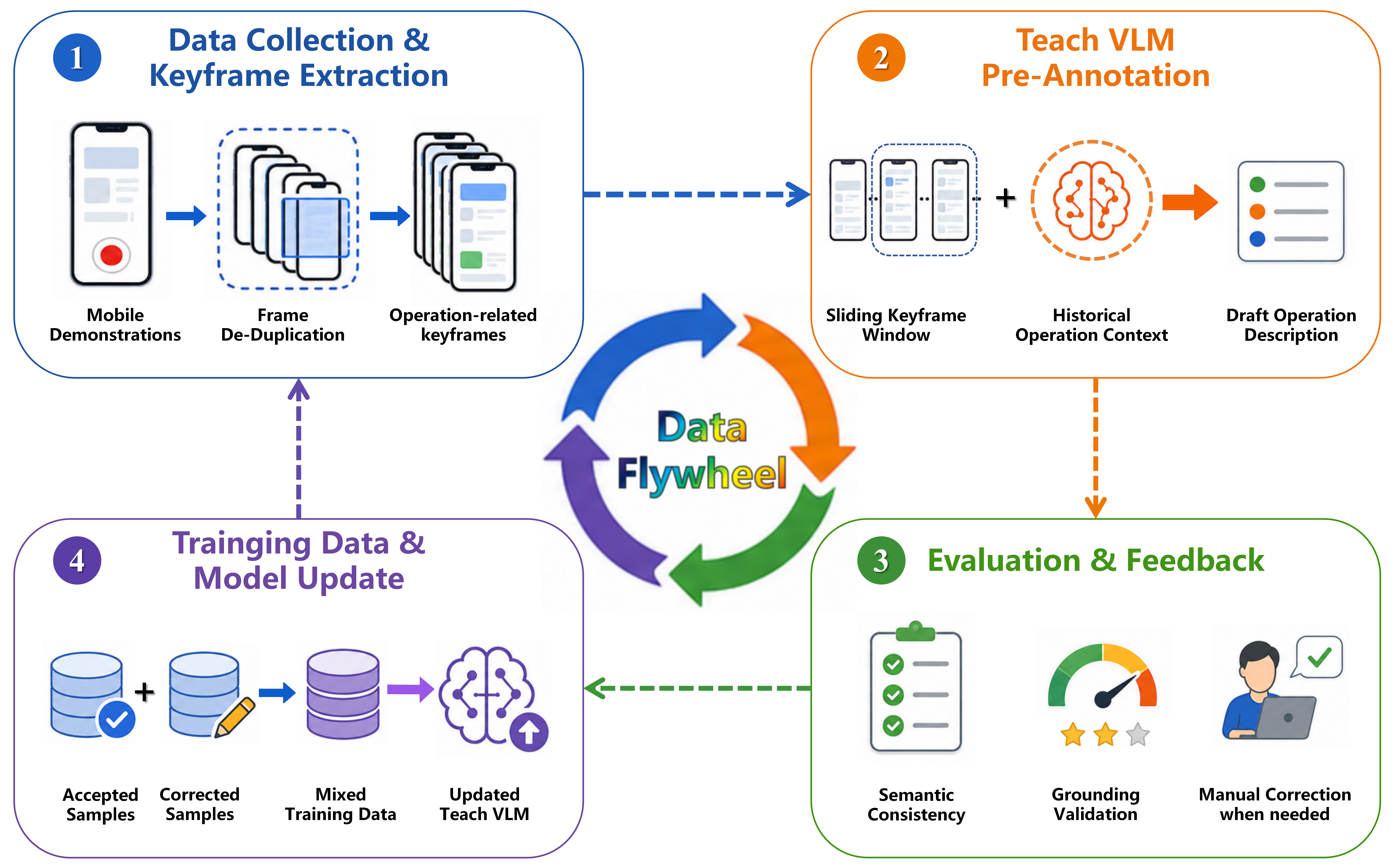} 
    \caption{Data flywheel for iterative Teach VLM improvement. The pipeline collects mobile demonstrations, extracts operation-related keyframes, uses Teach VLM to produce draft operation descriptions, refines the annotations through automatic evaluation and manual correction, and updates the model with accepted and corrected training data for the next iteration.}
    \label{fig:data_flywheel}
\end{figure*}

To extract accurate operational knowledge from noisy mobile demonstrations, we develop a data flywheel that iteratively constructs high-quality training data to fine-tune a general-purpose vision-language model into Teach VLM. As shown in Figure~\ref{fig:data_flywheel}, the flywheel cycles through five stages: (1) mobile demonstration collection, (2) operation-related keyframe extraction, (3) Teach VLM pre-annotation, (4) automatic evaluation with manual feedback, and (5) model retraining with the refined data. Each iteration improves both the model and the training corpus, progressively reducing the need for manual annotation. This data flywheel constitutes the core of our Teach phase, empowering the Teach VLM to generate high-quality procedural references that subsequently guide the downstream Repeat execution. The following subsections describe these stages in order.

\subsection{Data collection and key-frame extraction}

The acquisition of key-frame data related to user operations is the foundation for subsequently inferring operational knowledge. Irrelevant page-loading animations would impose unnecessary computational overhead on the model, and such noise would undermine the accuracy of the generated operational knowledge. To this end, we use a two-stage keyframe extraction pipeline. A rule-based script first scans densely sampled frames from a demonstration video and produces candidate before-and-after transitions according to visual change signals. A lightweight LLM then filters these candidates by judging whether the transition is likely to correspond to a meaningful user operation rather than loading, animation, or other operation-irrelevant visual changes.

\textbf{Extraction procedure.} The keyframe extraction module follows a compact four-stage pipeline:
\begin{enumerate}
    \item densely capture screenshots from a user demonstration at a fixed interval;
    \item compute and smooth frame-difference scores to remove visually redundant frames;
    \item segment stable and changing regions using stability and change thresholds;
    \item filter candidate before-and-after transitions with a lightweight LLM for operation relevance.
\end{enumerate}
This procedure preserves screen transitions likely to correspond to user operations while filtering repeated frames, loading animations, and other visually salient but operation-irrelevant changes. The keyframe filtering stage is independent from Teach VLM; Teach VLM is only used afterward for operation-description pre-annotation.

\textbf{Screenshot de-duplication.} To remove redundant frames, we design a frame-difference–based selection pipeline. Given a dense frame sequence $\{I_t\}_{t=1}^T$, where $I_t\in \mathbb{R}^{H\times W\times 3}$ denotes the RGB image at time $t$, we first convert each frame from RGB to the LUV color space. For each frame, we compute the absolute pixel-wise difference with respect to the previous frame and aggregate it by summation to obtain a scalar frame difference score. 
\begin{equation}
    d_t = \sum_{x,y,c} |L_t(x, y, c)-L_{t-1}(x,y,c)|
\end{equation}
where $t= 2, \dots, T$. This produces a 1D difference signal $\{d_t\}$ over time. To reduce local fluctuations, we apply a moving-average smoothing with window size $\omega$:
\begin{equation}
    \tilde{d}_t = \frac{1}{\omega}\sum_{k=t-\lfloor \omega/2\rfloor}^{t+\lfloor \omega/2\rfloor} d_k
\end{equation}
where boundary indices are clipped to $[2, T]$ in implementation.

We introduce two thresholds, a stability threshold $\tau_{stable}$ and a change threshold $\tau_{change}$, to distinguish stable regions from changing regions. Specifically, time steps with $\tilde{d}_t<\tau_{stable}$ are treated as visually stable, while time steps with  $\tilde{d}_t\geq\tau_{change}$ indicate significant visual changes. In the de-duplication stage, we traverse the sequence and remove long runs of stable frames. For each detected changing segment, we explicitly keep the last stable frame before the change, and the first frame inside the changing segment. This yields a compact set of candidates $\mathcal{K} = \{I_{t1}, I_{t2}, ..., I_{tM}\}$.

\textbf{Operation-related keyframe filtering.} Although the image de-duplication effectively removes most visually redundant screenshots, it is still agnostic to the \emph{cause} of the page transition. In particular, it cannot distinguish whether a visual change is triggered by a user operation (e.g., a click or swipe) or by autonomous in-app loading and animations. To bridge this gap, we use a lightweight LLM as a candidate-frame filter. Given a candidate before-and-after screenshot pair, the LLM judges whether the transition reflects a meaningful user operation and discards pairs dominated by splash screens, transient loading states, keyboard animations, or repeated end states. This design keeps keyframe selection separate from Teach VLM training and avoids using the target operation-description model as the keyframe selector.

\subsection{Pre-annotation}

To facilitate model iteration, we have built a semi-automatic annotation system. In the first iteration, the base Qwen3-VL-8B model without fine-tuning is used to produce initial pre-annotations. In subsequent iterations, the Teach VLM checkpoint from the previous round, denoted as $\theta_{f-1}$, serves as the pre-annotator. Specifically, $\theta_{f-1}$ pre-annotates newly collected demonstrations by inferring a natural-language operation description for each keyframe. These pre-annotations serve as inputs for automatic evaluation and manual feedback; they complement, rather than replace, human annotations and converted public datasets in the final training pool. During the inference process of Teach VLM, we exploit contextual information by combining a keyframe sliding window with historical operation descriptions, to extract operational knowledge embedded in continuous frame changes, and finally summarize natural-language operation descriptions. The inference formula is as follows:
\begin{equation}
\{D_j\}_{j=n}^{n+k-1} = \theta_{f-1}(\{I_i\}_{i=n}^{n+k}, \{D_i\}_{i=1}^{n-1})
\label{eq:inference_formula}
\end{equation}

where $\{I_i\}_{i=n}^{n+k}$ represents multiple keyframe screenshots within the sliding window, $\{D_i\}_{i=1}^{n-1}$ represents the step-by-step operation descriptions before the n-th frame, and the output $\{D_j\}_{j=n}^{n+k-1}$ is the corresponding natural-language operation descriptions within the interval. Inferring user operation semantics in the form of a sliding window can avoid distribution discrepancies caused by varying lengths of teach trajectories. It also constrains the output operation description steps to correspond one-to-one with the input keyframes, effectively preventing the hallucination of spurious or missing steps.

\subsection{Auto-evaluation and manual feedback}

We automatically evaluate the pre-annotated results in two ways: semantic assessment and grounding consistency assessment. For the semantic assessment, we leverage a powerful general-purpose VLM such as GPT-4o as the critic model to assess whether the description of the current step aligns with the transformations observed between consecutive frames, yielding a semantic score $c_s$ in the range 0 to 1. 

For grounding consistency assessment, given the predicted operation description and the corresponding screenshot, we use an open-source screen-grounded executor to predict the target element's coordinates. The intuition is that an accurate operation description should enable the executor to locate the correct target on the screen. We then compare the predicted coordinates with the ground-truth target location to obtain a grounding score $c_g$:
\begin{equation}
    c_g = e^{-\lambda \cdot d(P_{\text{g}}, P_{\text{gt}})}
\end{equation}

where $P_{\text{g}}$ and $P_{\text{gt}}$ denote the grounding coordinates predicted by the executor and the ground-truth target location, respectively. $\lambda$ is a hyper-parameter that controls the decay rate of the grounding score. $d(\cdot, \cdot)$ denotes the Euclidean distance between two vectors. This grounding score is used only for flywheel filtering and is not the primary output of the final Teach VLM.

Finally, we combine the two scores to obtain the final score $c_f$, and then filter high-confidence and low-confidence samples based on a threshold $\tau$:
\begin{equation}
    c_f = \alpha c_s + (1-\alpha)c_g 
\end{equation}
High-confidence samples are accepted into the training pool, while low-confidence or ambiguous samples are routed to manual correction before being reused for model update.

\subsection{Training}

\textbf{Data mixture.} A vision-language model for operational knowledge generation must recognize the target screen element, understand the page context, and infer the operation that links two or more consecutive screen states. Therefore, the primary supervision for Teach VLM is screen-transition data with natural-language operation descriptions. In our final training setting, this includes operation-description data derived from Android-In-The-Zoo, GUIOdyssey, AndroidControl, self-collected demonstrations, and self-constructed multi-image trajectories. Android-In-The-Zoo and GUIOdyssey provide before-and-after screenshots and multi-step mobile operation trajectories; AndroidControl and self-collected data improve target-element association and screen-context understanding; self-constructed multi-image trajectories further supplement localized and multi-frame scenarios. These data directly teach the model to map before-and-after screenshots or multi-keyframe sequences to step-wise operation descriptions, while operational knowledge generation remains the main training objective.

\textbf{Model update.} After pre-annotation and filtering, high-confidence samples and manually corrected samples are merged into the next training set. The updated Teach VLM is then used for the next round of pre-annotation, closing the data flywheel. This iterative update mechanism allows the system to improve both the model and the training corpus without treating any single automatic annotation round as final. The resulting fine-tuned model from this iterative process is Teach VLM, a Qwen3-VL-8B model specialized for operational knowledge generation from mobile screen transitions.

\section{Experiments}~\label{sec:experiments}
In this section, we evaluate Teach VLM as a screen-to-operational-knowledge model. We first detail the experimental setup, including implementation details, evaluation metrics, and datasets, with particular emphasis on our newly constructed Mobile Screen Teach Benchmark. The main evaluation focuses on whether Teach VLM can recover single-step and multi-step operational semantics from mobile screen transitions. We then examine whether the extracted knowledge remains useful for repeated task execution when consumed by existing execution models in dynamic Android environments. We further validate key design choices through ablation studies and qualitative case analysis.

\subsection{Implementation Details}

\textbf{Base Model:} We use the Qwen3-VL-8B-Instruct~\cite{qwen3vltechnicalreport,qwen3vl8binstruct} open-source vision-language model as the foundation model to train our Teach VLM. This model possesses general visual understanding and screen-grounding capabilities.

\textbf{Training Setup:} We implement our training framework using the SWIFT library~\cite{zhao2024swiftascalablelightweightinfrastructure}. The base model is fine-tuned using LoRA (Low-Rank Adaptation) to ensure parameter efficiency. Specifically, we set the LoRA rank to $8$, alpha to $32$, and apply it to all linear layers. The model is trained for $3$ epochs with a maximum sequence length of $12288$; over-length samples are skipped and resampled by the training framework. We set the peak learning rate to $1 \times 10^{-4}$ with a warmup ratio of $0.05$. To optimize memory usage and accelerate training, we employ DeepSpeed ZeRO-2 and Flash Attention, and conduct the entire training process in bfloat16 precision. The training is distributed across $8$ GPUs, with a per-device batch size of $2$ and a gradient accumulation step of $2$, resulting in an effective global batch size of $32$. For visual inputs, we constrain the image resolution with a maximum of $1,003,520$ pixels and a minimum of $3,136$ pixels to balance computational cost and visual detail preservation. Unless otherwise stated, all Teach VLM results use the best validation checkpoint at step $5200$.

\subsection{Datasets}
We conduct experiments on four representative data sources and benchmarks to evaluate operational knowledge extraction and downstream utility.

\textbf{Android-In-The-Zoo (AiTZ)}~\cite{ZhangWTLXXWT24} is a large-scale dataset containing 18,643 screen-action pairs and over 2,500 instructions across diverse Android applications. It provides semantic annotations, including action descriptions and execution results, alongside paired screenshots depicting the before and after states of individual operations. This structure supports evaluation of both single-step operation description prediction and multi-step operation sequence understanding.

\textbf{GUIOdyssey}~\cite{lu2025guiodyssey} is a comprehensive benchmark for \textbf{cross-app} mobile interaction understanding. It consists of over 8,300 human-demonstrated episodes spanning 212 diverse applications and 1.4K app combinations. The dataset covers diverse interaction patterns across multiple devices and task categories, making it suitable for testing whether operational knowledge extraction generalizes to cross-app screen transitions.

\textbf{Android World Benchmark}~\cite{rawles2024androidworlddynamicbenchmarkingenvironment} is a highly reproducible and dynamic benchmark for evaluating screen-based task execution on a live Android emulator. It features 116 hand-crafted tasks across 20 real-world applications, which are dynamically instantiated with randomly-generated parameters to create millions of unique task variations. The benchmark provides task-specific reward checkers and built-in difficulty labels, with 61 easy, 36 medium, and 19 hard task templates in our task list. We use this benchmark as an external testbed for assessing the downstream utility of Teach-generated operational knowledge in complex, dynamic environments.

\textbf{Mobile Screen Teach Benchmark} is a self-constructed video-based dataset designed for evaluating operational knowledge extraction from demonstration videos. It focuses on Chinese-localized Android applications and contains screen operation videos recorded from popular real-world apps across domains such as food delivery, express delivery, and e-commerce. All operation descriptions and annotations are natively in Chinese, providing a testbed for non-English mobile screen understanding. Each video is annotated with frame-level keyframes and corresponding natural-language operation descriptions, enabling evaluation of both single-step operation understanding and multi-step operation sequence extraction. In our test split, the keyframe-based evaluation protocol contains 169 Chinese task trajectories with 1,250 annotated operation steps. We additionally evaluate a fixed-interval video-frame protocol, which contains 131 screen-recording trajectories with 888 annotated operation steps, to test robustness under noisier frame sampling. This benchmark bridges noisy video demonstrations and operational knowledge evaluation in localized mobile environments.

\subsection{Evaluation Metrics}~\label{sec:eval_metrics}

We evaluate our method from two aspects: operational knowledge extraction quality and downstream utility.

\paragraph{Operational Knowledge Extraction Quality.}
To evaluate the quality of extracted operational knowledge, we convert each natural language operation description into a structured tuple $(action, target, argument)$ using rule-based parsing. Two operations are considered matched if the action types are identical and the target elements are semantically similar. For target-element matching, we first normalize both predicted and ground-truth element strings by lowercasing text, removing punctuation and quotes, replacing underscores or hyphens with spaces, and dropping common articles such as ``the'', ``a'', and ``an''. The normalized targets are then matched if they are identical, if one token set is contained in the other, or if their token-level Jaccard overlap is at least 0.6. This soft matching allows minor surface-form variation, such as \texttt{search icon} versus \texttt{Search}, while still requiring substantial lexical overlap. Element and argument metrics are computed only on samples where the corresponding slot is applicable, whereas OSA checks whether all applicable slots of an operation are correct.

\textbf{Single-step Operational Knowledge.}
For single-step operation understanding, the input consists of two screenshots representing the screen states before and after an operation, and the model predicts the corresponding operation description. We evaluate the predictions using four metrics.
Action Accuracy (AA) measures whether the predicted action type matches the ground-truth action.
Element Matching Accuracy (EMA) evaluates whether the predicted target screen element corresponds to the ground-truth element.
Argument Accuracy (ArA) measures whether the predicted argument matches the ground-truth argument.
Operation Semantic Accuracy (OSA) measures whether the complete operation semantics are correctly predicted, which requires the action type, the target screen element, and the argument to be correctly identified.

\textbf{Multi-step Operational Knowledge.}
For multi-step operation understanding, the model generates a sequence of operation descriptions. We evaluate the predicted sequences using step-level Precision, Recall, and F1-score based on the matching between predicted and ground-truth operations. In addition, we introduce the Redundant Action Rate (RAR) to measure the proportion of redundant operations generated by the model:
$$
RAR = \frac{N_{redundant}}{N_{predicted}}
$$
where $N_{predicted}$ is the total number of predicted operation steps, and $N_{redundant}$ denotes the number of predicted steps that do not match any ground-truth operation under the matching rule defined above.

\paragraph{Downstream Utility.}
To measure whether the operational knowledge summarized by our method remains useful beyond offline screen understanding, we adopt Task Success Rate (SR) as the primary metric in repeated execution experiments.

Task Success Rate (SR) measures the proportion of tasks successfully completed by the execution system:
$$
SR = \frac{N_{success}}{N_{total}}
$$
where $N_{success}$ and $N_{total}$ denote the number of successful tasks and the total number of tasks, respectively.
For answer-style tasks, we count a trajectory as successful if it produces the correct answer but only misses the final termination call. Repeated-answer-with-finish failures are not corrected, since they may reflect wrong answers, incorrect termination timing, or reward-checker mismatch.

\subsection{Main Results}
\subsubsection{Teach VLM Evaluation}

\begin{table*}[t]
\centering
\caption{Performance comparison on single-step operation description prediction. Baseline references: GPT-4o~\cite{openai2024gpt4osystemcard}, Doubao-Seed-1.8~\cite{bytedanceseed2026seed18}, Doubao-Seed-2.0~\cite{bytedanceseed2026seed20}, Qwen3.5-9B~\cite{qwen2026qwen35}, and Qwen3-VL-8B~\cite{qwen3vltechnicalreport,qwen3vl8binstruct}.}
\label{tab:teach_single_step}
\begin{tabular}{lccccc}
\hline
Model & Dataset & AA ($\uparrow$) & EMA ($\uparrow$) & ArA ($\uparrow$) & OSA ($\uparrow$) \\
\hline
GPT-4o & Android-In-The-Zoo & 20.93 & 23.60& 12.70 & 16.07\\
Doubao-Seed-1.8 & Android-In-The-Zoo & 61.66 & 35.37 & 56.85 & 30.28 \\
Doubao-Seed-2.0 & Android-In-The-Zoo & 64.08 & 37.04 & 42.37 & 42.37 \\
Qwen3.5-9B & Android-In-The-Zoo & 22.71 & 25.55 & 29.81 & 17.38 \\
Qwen3-VL-8B & Android-In-The-Zoo & 41.35 & 27.19 & 57.22 & 23.07 \\
\textbf{Teach VLM} & Android-In-The-Zoo & \underline{\textbf{82.10}} & \underline{\textbf{67.66}} & \underline{\textbf{92.91}} & \underline{\textbf{58.84}}\\
\hline
GPT-4o & GUIOdyssey & 11.48 & 40.14 & 3.35 & 2.12 \\
Doubao-Seed-1.8 & GUIOdyssey & 67.15 & 32.07 & 34.32 & 13.97 \\
Doubao-Seed-2.0 & GUIOdyssey & 66.63 & 31.25 & 43.43 & 16.55 \\
Qwen3.5-9B & GUIOdyssey & 32.59 & 40.96 & 26.14 & 11.90 \\
Qwen3-VL-8B & GUIOdyssey & 34.61 & 36.42 & 30.03 & 9.57 \\
\textbf{Teach VLM} & GUIOdyssey & \underline{\textbf{78.22}} & \underline{\textbf{45.55}} & \underline{\textbf{78.28}} & \underline{\textbf{42.99}} \\
\hline
\end{tabular}
\end{table*}

\begin{table*}[t]
\centering
\caption{Performance comparison on multi-step operation sequence prediction. Baseline references follow Table~\ref{tab:teach_single_step}.}
\label{tab:teach_multi_step}
\begin{tabular}{lccccc}
\hline
Model & Dataset & Precision ($\uparrow$) & Recall ($\uparrow$) & F1-score ($\uparrow$) & RAR ($\downarrow$) \\
\hline
GPT-4o & Android-In-The-Zoo & 19.15 & 16.70 & 17.84 & 80.85 \\
Doubao-Seed-1.8 & Android-In-The-Zoo & 44.13 & 35.56 & 39.39 & 55.87 \\
Doubao-Seed-2.0 & Android-In-The-Zoo & 41.14 & 36.03 & 38.42 & 58.86 \\
Qwen3.5-9B & Android-In-The-Zoo & 5.79 & 13.27 & 8.06 & 94.21 \\
Qwen3-VL-8B & Android-In-The-Zoo & 34.62 & 25.70 & 29.50 & 65.38 \\
Teach VLM & Android-In-The-Zoo & \underline{\textbf{74.43}} & \underline{\textbf{73.85}} & \underline{\textbf{74.14}} & \underline{\textbf{25.56}} \\
\hline
GPT-4o & GUIOdyssey & 18.46 & 16.69 & 17.53 & 81.54 \\
Doubao-Seed-1.8 & GUIOdyssey & 27.46 & 25.46 & 26.42 & 72.54 \\
Doubao-Seed-2.0 & GUIOdyssey & 23.90 & 22.46 & 23.16 & 76.10 \\
Qwen3.5-9B & GUIOdyssey & 5.18 & 8.77 & 6.51 & 94.82 \\
Qwen3-VL-8B & GUIOdyssey & 26.40 & 20.03 & 22.78 & 73.60 \\
Teach VLM & GUIOdyssey & \underline{\textbf{46.18}} & \underline{\textbf{46.78}} & \underline{\textbf{46.48}} & \underline{\textbf{53.81}} \\
\hline
\end{tabular}
\end{table*}

\begin{table*}[t]
\centering
\caption{Performance comparison on video-based multi-step operational knowledge extraction on Mobile Screen Teach Benchmark. Baseline references follow Table~\ref{tab:teach_single_step}.}
\label{tab:video_multi_step}
\begin{tabular}{lcccc}
\hline
Model & Precision ($\uparrow$) & Recall ($\uparrow$) & F1-score ($\uparrow$) & RAR ($\downarrow$) \\
\hline
GPT-4o & 30.45 & 33.70 & 32.00 & 69.55 \\
Doubao-Seed-2.0 & 45.25 & 49.28 & 47.18 & 54.75 \\
Doubao-Seed-1.8 & 37.42 & 41.38 & 39.30 & 62.58 \\
Qwen3.5-9B & 1.70 & 20.91 & 3.15 & 98.30 \\
Qwen3-VL-8B & 37.53 & 37.15 & 37.34 & 62.47 \\
Teach VLM & \underline{\textbf{56.96}} & \underline{\textbf{53.68}} & \underline{\textbf{55.27}} & \underline{\textbf{43.03}} \\
\hline
\end{tabular}
\end{table*}
\textbf{Single-step Operation Prediction.} To evaluate the model's capability in fine-grained target-element grounding and atomic action recognition, we first analyze single-step operation prediction. As shown in Table \ref{tab:teach_single_step}, Teach VLM consistently outperforms all baselines on both Android-In-The-Zoo and GUIOdyssey, achieving the highest scores across all four metrics on both datasets. In particular, on the strictest joint-correctness metric (OSA), Teach VLM reaches 58.84\% and 42.99\% on Android-In-The-Zoo and GUIOdyssey respectively, outperforming the strongest baseline (Doubao-Seed-2.0) by 16.47 and 26.44 percentage points.

Notably, general-purpose VLMs exhibit a pronounced multi-element joint prediction weakness on this task. For instance, although Qwen3.5-9B achieves a moderate EMA (40.96\%) on GUIOdyssey, its ArA and OSA fall to 26.14\% and 11.90\%, indicating that detecting screen elements alone is far from sufficient for complete operation understanding. In particular, GPT-4o scores particularly low under the rule-based parsing protocol defined in Section~\ref{sec:eval_metrics} (OSA of only 2.12\% on GUIOdyssey), primarily because it tends to generate free-form descriptions that are hard to parse into the structured operational tuple. In contrast, Teach VLM, fine-tuned through the data flywheel, accurately predicts a coherent and accurate overall operational semantics that jointly covers action, target, and argument.

\textbf{Multi-step Operation Prediction.} Beyond atomic actions, real-world mobile tasks involve sequences of operations. We evaluate Teach VLM on multi-step prediction under two settings: (1) predicting operational knowledge sequences from multi-frame screen-state sequences in Android-In-The-Zoo and GUIOdyssey, and (2) deriving operational knowledge sequences from demonstration videos in Mobile Screen Teach Benchmark.

As shown in Table \ref{tab:teach_multi_step} and Table \ref{tab:video_multi_step}, Teach VLM maintains high accuracy on complex multi-step sequences while exhibiting strong hallucination resistance. On Android-In-The-Zoo, Teach VLM achieves the highest F1-score of 74.14\% with a RAR of only 25.56\%. In contrast, GPT-4o and Qwen3.5-9B suffer RAR above 80\%, indicating that general-purpose VLMs tend to fabricate a large number of spurious intermediate steps when handling sequence prediction.

This advantage is further amplified in more challenging settings. On GUIOdyssey, which involves cross-app navigation, Teach VLM improves the best baseline F1 (Doubao-Seed-1.8, 26.42\%) to 46.48\%. On Mobile Screen Teach Benchmark, where uniformly sampled video frames introduce temporal noise, Teach VLM outperforms Doubao-Seed-2.0 by 8.09 percentage points. These results strongly indicate that Teach VLM accurately captures the causal transition logic between consecutive screen states. Given a keyframe sequence, it not only precisely infers the real operations that cause screen-state changes, but also strictly follows the actual interaction process, effectively avoiding the action hallucination problem that general VLMs are prone to in multi-step reasoning.

\subsubsection{Downstream Utility of Operational Knowledge}

\begin{table*}
\centering
\small
\setlength{\tabcolsep}{4pt}
\begin{tabular}{lccccc}
\toprule
\textbf{Model} & \textbf{Setting} & \multicolumn{4}{c}{\textbf{Task Success Rate ($\uparrow$)}} \\
\cmidrule(lr){3-6}
 & & Easy & Medium & Hard & Overall \\
\midrule
\multirow{2}{*}{Qwen3-VL-8B}
& w/o Teach VLM & 36.06 & 11.11 & 7.89 & 23.70 \\
& w/ Teach VLM & 42.62 (+6.56) & 19.44 (+8.33) & 15.79 (+7.90) & 31.03 (+7.33) \\
\midrule
\multirow{2}{*}{GPT-4o}
& w/o Teach VLM & 52.45 & 16.66 & 18.42 & 35.77 \\
& w/ Teach VLM & 63.93 (+11.48) & 30.56 (+13.90) & 21.05 (+2.63) & 46.55 (+10.78) \\
\midrule
\multirow{2}{*}{Doubao-Seed-2.0}
& w/o Teach VLM & 65.57 & 36.11 & 21.05 & 49.14 \\
& w/ Teach VLM & 80.33 (+14.75) & 44.44 (+8.33) & 26.32 (+5.26) & 60.34 (+11.21) \\
\bottomrule
\end{tabular}
\caption{\label{tab:teach_vlm_execution_performance}
External utility evaluation of Teach-generated operational knowledge on the Android World benchmark. We report Task Success Rate over 61 Easy, 36 Medium, and 19 Hard task templates; for answer-style tasks, correct answers without the final finish call are counted as successful. ``w/o Teach VLM'' denotes the zero-shot executor baseline; ``w/ Teach VLM'' denotes the same executor augmented with Teach-injected operational knowledge as an execution reference.
Execution backbone references: Qwen3-VL-8B~\cite{qwen3vltechnicalreport,qwen3vl8binstruct}, GPT-4o~\cite{openai2024gpt4osystemcard}, and Doubao-Seed-2.0~\cite{bytedanceseed2026seed20}.
}
\end{table*}

The primary goal of Teach VLM is to extract natural-language operational knowledge from screen demonstrations; we now verify that this knowledge remains useful when reused by existing execution models. The evaluation is conducted on Android World, which features tasks of varying difficulty and real app states.

Specifically, for each test task we first collect a human demonstration trajectory on the Android World emulator and use Teach VLM to extract operational knowledge. We then attach this knowledge to the executor's system prompt as an \textit{execution reference} (e.g., ``Reference operation sequence: 1. tap Search $\rightarrow$ 2. enter query $\rightarrow$ ...''), which is consumed alongside the task instruction at every decision step. This does not change the underlying executor or its action space, but only provides an external guide of the intended procedure. We evaluate three execution backbones (GPT-4o, Qwen3-VL-8B, and Doubao-Seed-2.0) with different screen-understanding capabilities. This step corresponds to the \textit{repeat} phase, which reuses extracted operation descriptions as procedural references rather than replaying recorded actions.

As shown in Table \ref{tab:teach_vlm_execution_performance}, the operational knowledge extracted by Teach VLM brings substantial gains across all three backbones. Qwen3-VL-8B, GPT-4o, and Doubao-Seed-2.0 obtain +7.33, +10.78, and +11.21 percentage points overall, respectively. The gains are most pronounced on Medium-difficulty tasks (e.g., GPT-4o +13.90 pp), which typically involve complex cross-page navigation; under the zero-shot setting (w/o Teach), executors often drift into ineffective exploration loops, while the trajectory description provided by Teach VLM serves as a demonstration-derived reference path that directly eliminates such loops. On Hard tasks, even when Teach VLM effectively resolves the high-level planning question of what to do next, executors still fail because of low-level execution errors, including imprecise element grounding, fragile multi-step state tracking, and incorrect text input.

These results strongly indicate that the descriptions generated by Teach VLM encode valuable procedural information that can substantially reduce task-planning difficulty. At the same time, the fact that overall SR does not reach 100\% points to a clear direction for future screen-based execution systems: once high-level procedural knowledge is available, the community still needs to strengthen low-level capabilities of vision-language models, including grounding, long-horizon state tracking, and robust text input, to close the remaining execution gap.

\subsection{Ablation Studies}
\label{sec:ablation}

We ablate two key design choices of Teach VLM at sequence level: keyframe extraction and sliding-window inference. These ablations are conducted on Mobile Screen Teach Benchmark because it is the only dataset that provides both keyframe-aligned and raw video-frame inputs, allowing controlled comparison of the two design choices. The ablations are intended to validate the input construction and inference protocol rather than introduce a new model variant.

\begin{figure}[t]
\centering
\includegraphics[width=\columnwidth]{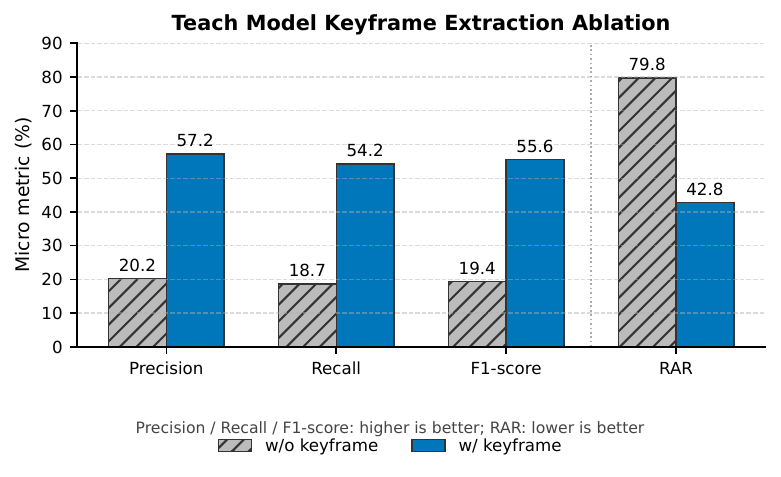}
\caption{\label{fig:keyframe_extraction_ablation}
Teach VLM ablation of keyframe extraction on sequence-level operational knowledge generation. Keyframe extraction improves micro precision, recall, and F1-score while substantially reducing the redundant action rate.}
\end{figure}

\textbf{Effect of keyframe extraction.} We compare operation-aligned key screenshots with non-keyframe video-frame inputs under the same sequence evaluation protocol. As shown in Figure \ref{fig:keyframe_extraction_ablation}, keyframe extraction is critical for reliable sequence prediction with Teach VLM. With keyframe extraction, Teach VLM achieves 57.18\% micro precision, 54.16\% micro recall, and 55.63\% micro F1, while keeping RAR at 42.82\%. Without keyframe extraction, performance drops to 20.24\% precision, 18.69\% recall, and 19.44\% F1, while RAR increases to 79.76\%. These results indicate that raw video-frame inputs contain many transition or redundant frames that are not aligned with discrete operations, which makes the model more likely to hallucinate redundant steps or miss true action boundaries.

\begin{figure}[t]
\centering
\includegraphics[width=\columnwidth]{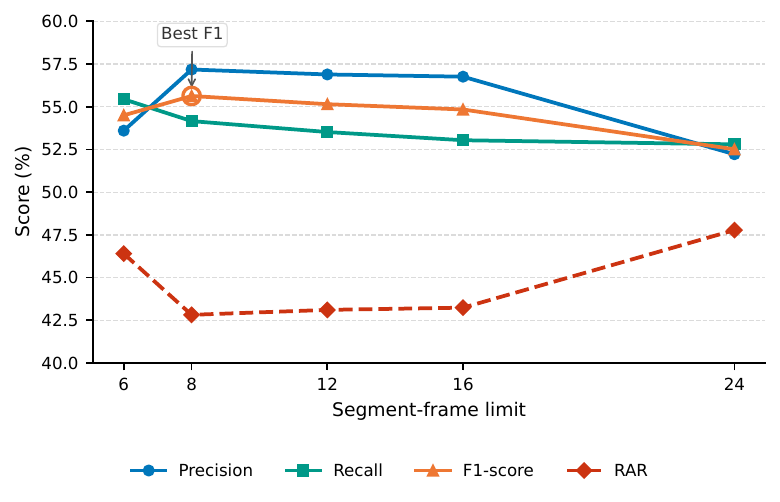}
\caption{\label{fig:window_ablation}
Ablation of inference window size on ground-truth-keyframe sequence prediction. The segment overlap is fixed to 2 frames. A moderate 8-frame window achieves the best F1-score and the lowest redundant action rate.}
\end{figure}

\textbf{Effect of inference window size.} We also vary the maximum number of frames in each inference segment while keeping the model, prompt, test-set samples, metric, and overlap size fixed. Figure \ref{fig:window_ablation} shows a trade-off between local action recall and sequence-level redundancy. A 6-frame window achieves the highest recall (55.44\%), but also produces more redundant predictions (RAR 46.40\%). The 8-frame window provides the best overall balance, reaching the highest F1-score (55.63\%), the highest precision (57.18\%), and the lowest RAR (42.82\%). Larger windows do not further improve performance, suggesting that moderate-length windows provide sufficient procedural context while avoiding the reasoning burden and redundancy introduced by overly long visual sequences.

\subsection{Case Study}
\label{sec:case_study}

\begin{figure*}[p]
\centering
\includegraphics[height=0.82\textheight,keepaspectratio]{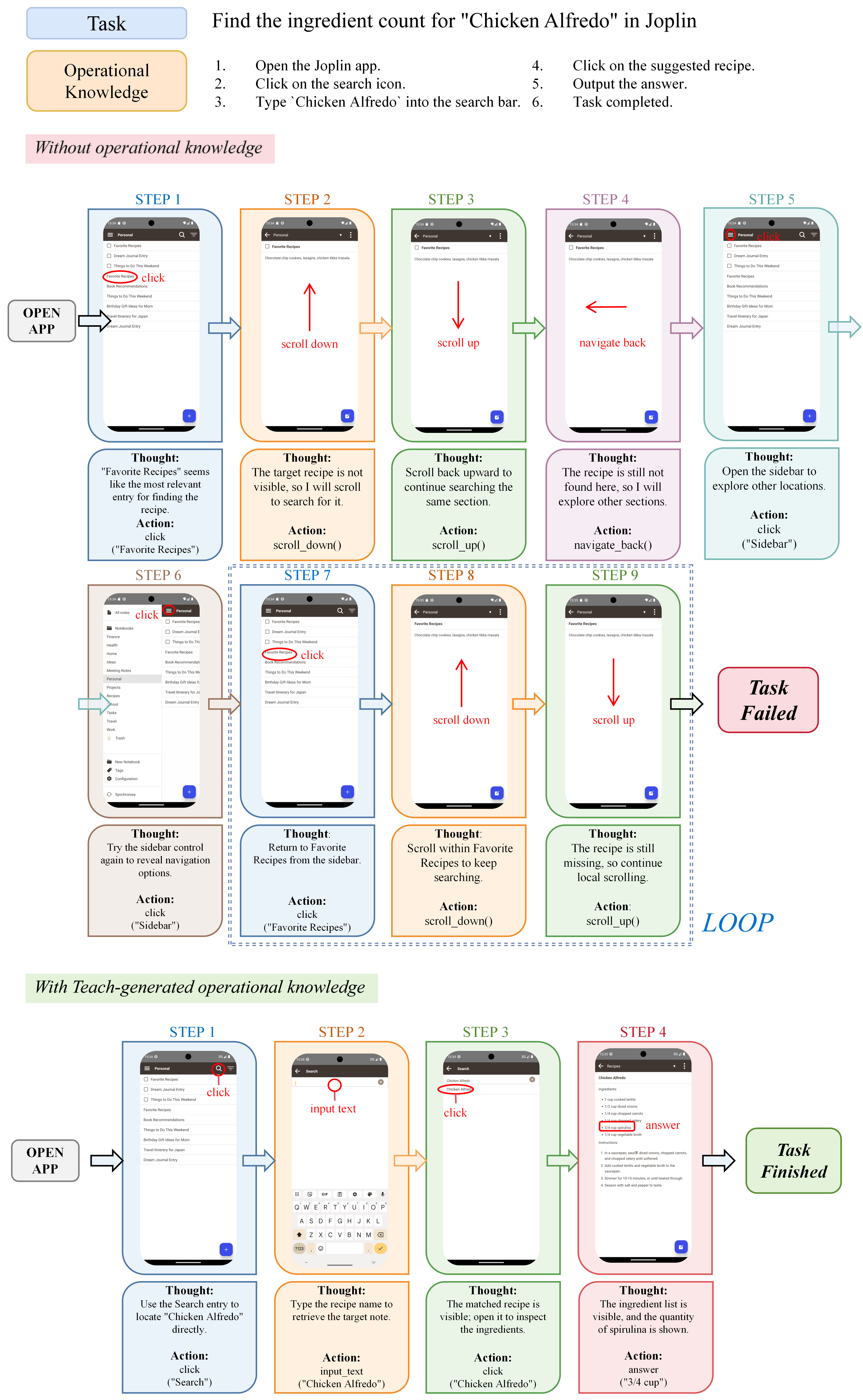}
\caption{\label{fig:case_notes_recipe}
Trajectory comparison on \texttt{NotesRecipeIngredientCount}. Operational knowledge guides the executor to use search and retrieve the target recipe, while the failure trajectory enters a browsing loop through \texttt{Favorite Recipes}.}
\end{figure*}

\begin{figure*}[p]
\centering
\includegraphics[height=0.82\textheight,keepaspectratio]{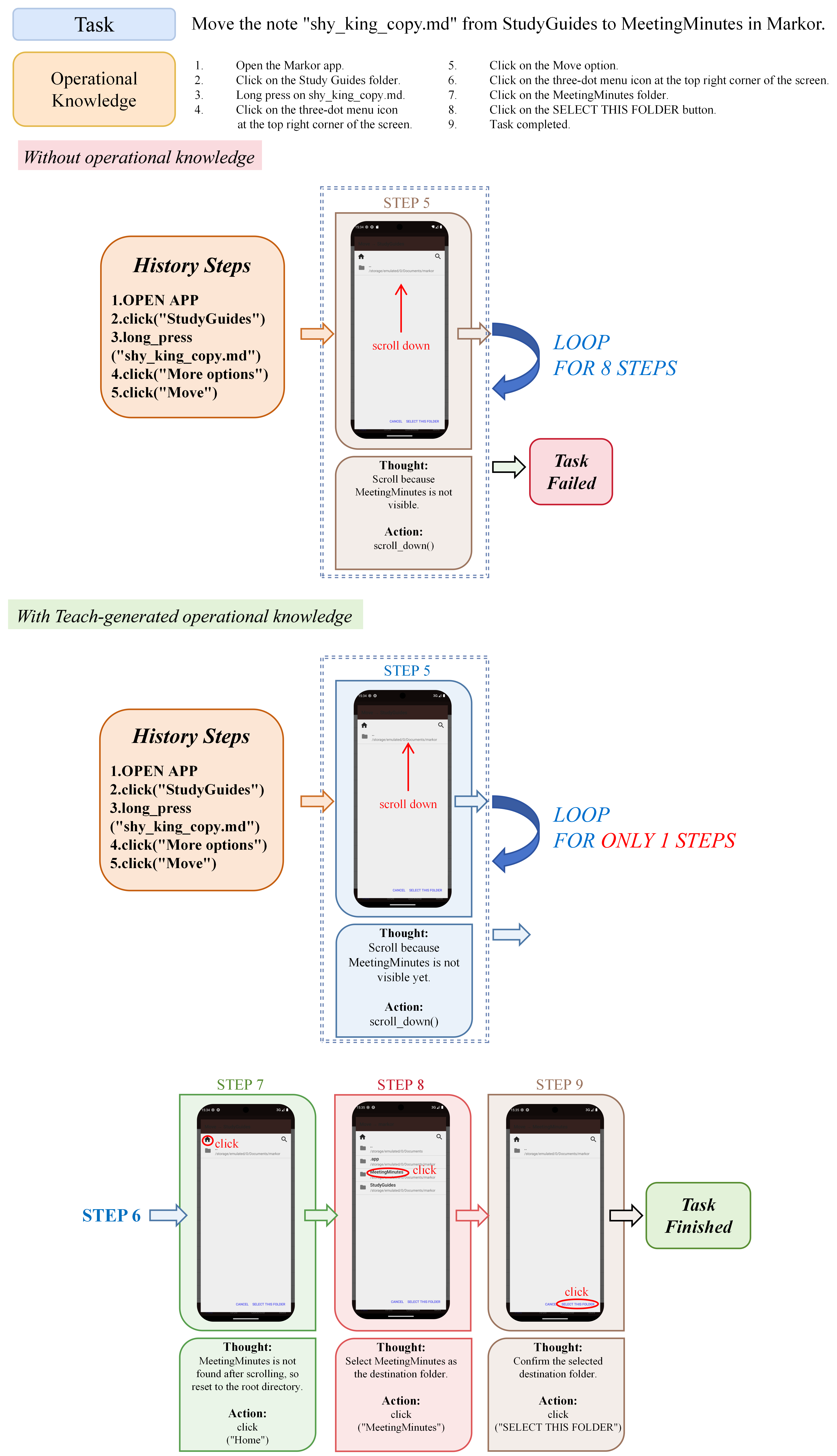}
\caption{\label{fig:case_markor_move}
Trajectory comparison on \texttt{MarkorMoveNote}. After the shared \texttt{Move} prefix, operational knowledge helps the executor switch navigation context and select \texttt{MeetingMinutes}, while the failure trajectory repeatedly scrolls within the same dialog.}
\end{figure*}

To better understand what kind of procedural information Teach VLM contributes, we examine paired trajectories from Android World with and without Teach-injected operational knowledge, selected to cover both successful and failed cases as well as different task types (retrieval, navigation, cross-app transcription, and numerical reasoning). The comparison, visualized in Figures \ref{fig:case_notes_recipe} and \ref{fig:case_markor_move}, illustrates how the extracted knowledge provides task-relevant intermediate guidance and reshapes the trajectory at key decision points.

\textbf{Success Case 1: NotesRecipeIngredientCount.} The system must find the amount of spirulina in the \texttt{Chicken Alfredo} recipe within Joplin. With operational knowledge, the successful trajectory uses \texttt{Search}, retrieves the matched note, and answers \texttt{3/4 cup}. The failure trajectory chooses \texttt{Favorite Recipes}, then loops through scrolling, backtracking, sidebar exploration, and re-entering the same section without reaching the target note. This case indicates that injected knowledge helps map a high-level retrieval goal to the correct functional entry point.

\textbf{Success Case 2: MarkorMoveNote.} This task requires moving a note from \texttt{StudyGuides} to \texttt{MeetingMinutes}. Both trajectories share the same prefix up to the \texttt{Move} stage. After that, the successful trajectory changes navigation context and reaches \texttt{MeetingMinutes}, while the failed trajectory remains in the same folder dialog and repeatedly scrolls. This case shows that Teach-generated operational knowledge also acts as a stage-level planning prior, constraining when the executor should continue local search and when it should switch context.

Taken together, these cases reveal two recurring forms of procedural screen semantics captured by Teach VLM. First, operational knowledge improves key entry-point selection, mapping high-level retrieval goals to correct functional entry points, especially for retrieval-style tasks where \texttt{Search} should be preferred over manual browsing. Second, it improves long-horizon trajectory stability by providing explicit intermediate goals. This reduces drift into repetitive scrolling or ineffective navigation loops observed in the failure trajectories of both cases. These qualitative findings complement Table \ref{tab:teach_vlm_execution_performance} by revealing the mechanism behind the SR gains: operational knowledge reshapes trajectories at decision points where executors otherwise drift into ineffective loops.

\begin{figure*}[p]
\centering
\includegraphics[height=0.90\textheight,keepaspectratio]{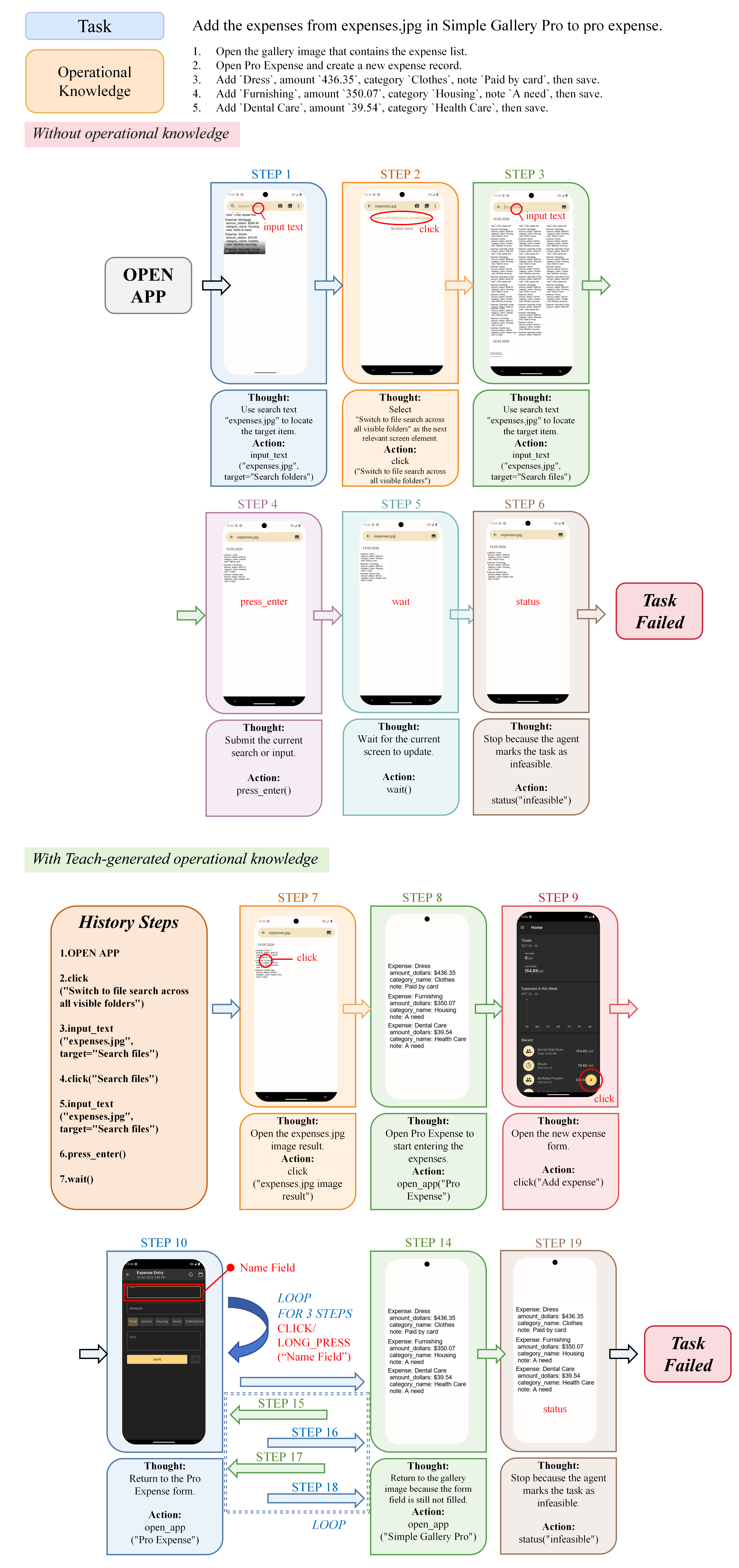}
\caption{\label{fig:case_expense_add_failure}
Trajectory comparison on \texttt{ExpenseAddMultipleFromGallery}. Operational knowledge guides the executor from gallery search to the Pro Expense entry form, but the task still fails because the executor cannot reliably transcribe multi-field expense information from the image into the form.}
\end{figure*}

\begin{figure*}[p]
\centering
\includegraphics[height=0.90\textheight,keepaspectratio]{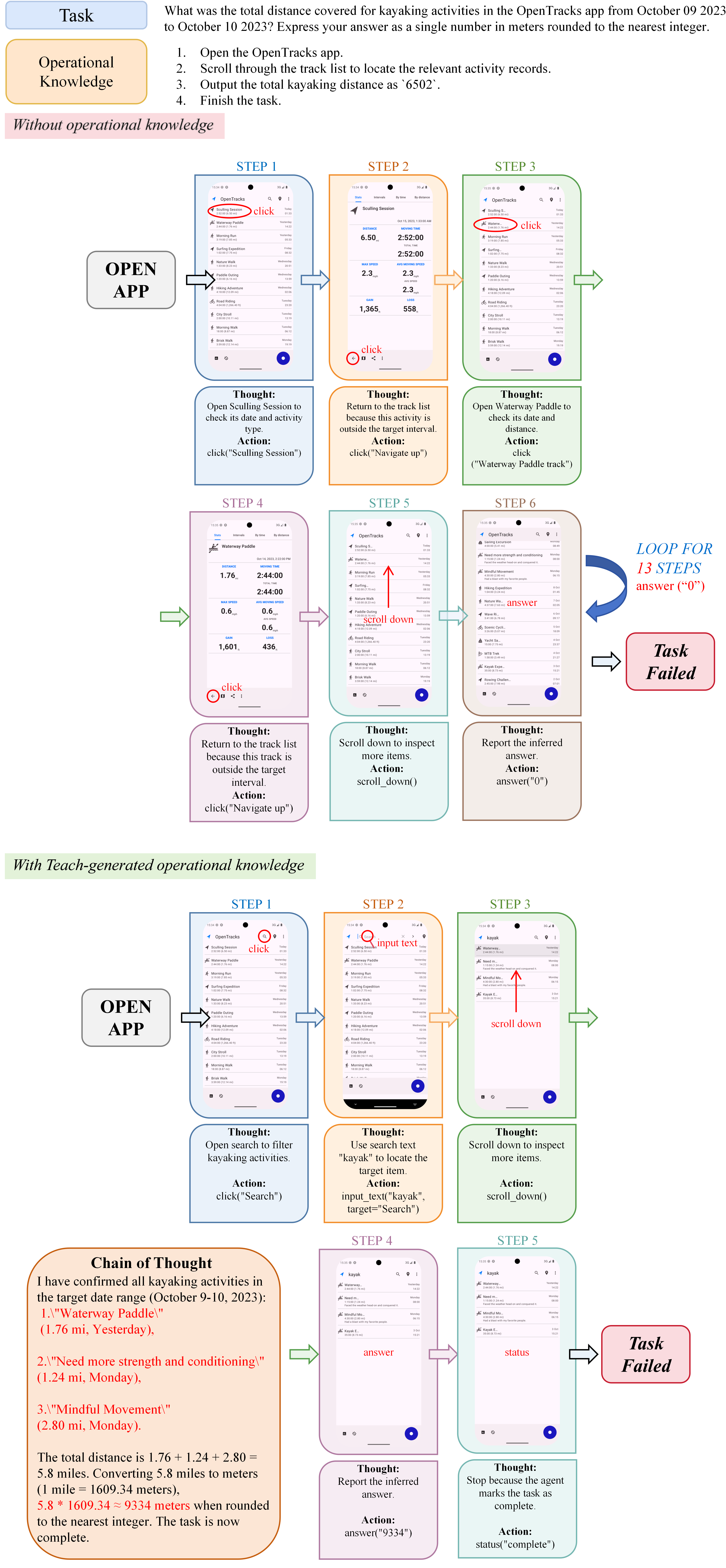}
\caption{\label{fig:case_sports_distance_failure}
Trajectory comparison on \texttt{SportsTrackerTotalDistanceForCategoryOverInterval}. Operational knowledge changes the trajectory from manual browsing to targeted search, but the executor still produces an incorrect answer due to temporal filtering and distance aggregation errors.}
\end{figure*}

We further examine two hard cases (both labeled as Hard in Android World), visualized in Figures \ref{fig:case_expense_add_failure} and \ref{fig:case_sports_distance_failure}, where operational knowledge changes the trajectory but does not fully solve the task, revealing the remaining execution bottlenecks after high-level procedural guidance is provided.

\textbf{Failure Case 1: ExpenseAddMultipleFromGallery.} This cross-app task requires selecting an expense image from the gallery and filling in multiple expense fields in the Pro Expense app. As shown in Figure \ref{fig:case_expense_add_failure}, the trajectory without operational knowledge remains trapped in the gallery-search stage. With operational knowledge, the executor reaches the target image and the Pro Expense entry form, but still fails to transcribe the multi-field expense records into the form. This case shows that Teach-generated knowledge can recover the intended cross-app procedure, while robust visual transcription and form filling remain necessary for final success.

\textbf{Failure Case 2: SportsTrackerTotalDistanceForCategoryOverInterval.} This task requires computing the total distance of activities in a given category over a specified time interval. As shown in Figure \ref{fig:case_sports_distance_failure}, the trajectory without operational knowledge manually inspects visible tracks, scrolls through the list, and repeatedly answers \texttt{0} (a wrong answer reflecting the absence of any aggregated distance). With operational knowledge, the executor instead opens \texttt{Search}, filters activities with \texttt{kayak}, and reasons over the returned candidates, but still outputs \texttt{9334} rather than the reference answer \texttt{6502}, indicating an error in distance aggregation across filtered activities. This case indicates that operational knowledge improves the retrieval strategy, while hard answer-style tasks still require accurate temporal filtering, distance aggregation, and answer verification.

These failure cases expose two complementary bottlenecks that operational knowledge alone cannot resolve. First, visual perception, including robust transcription and form filling, as shown in Figure \ref{fig:case_expense_add_failure}. Second, numerical reasoning, including temporal filtering, distance aggregation, and answer verification, as shown in Figure \ref{fig:case_sports_distance_failure}. Together, they indicate that Teach VLM primarily reduces uncertainty in step-level task progression, while final success on hard dynamic tasks still depends on complementary low-level execution abilities such as visual detail extraction, cross-screen memory, temporal filtering, and numerical verification.

\section{Conclusion}~\label{sec:conclusion}

In this paper, we tackle the challenge of extracting accurate operational knowledge from noisy mobile screen demonstrations. To this end, we introduce Teach VLM, a core model trained through a systematic data flywheel that enables scalable, low-cost construction of high-quality supervision. By filtering operation-irrelevant frames and modeling the causal transition logic between consecutive screen states, Teach VLM substantially mitigates the action hallucination problem that general vision-language models suffer from in multi-step reasoning, achieving state-of-the-art results on Android-In-The-Zoo, GUIOdyssey, and the newly released Chinese Mobile Screen Teach Benchmark. Building on this capability, we further propose the \textit{Teach-and-Repeat} paradigm, which decouples one-time teaching from repeated execution by converting unconstrained human demonstrations into inspectable, natural-language procedural references for downstream screen-based execution agents. On Android World, injecting these descriptions as an external execution reference yields consistent Task Success Rate improvements of +7.33 to +11.21 percentage points across three different execution backbones. Together, our results indicate that demonstration-derived operational knowledge serves as an effective and interpretable intermediate layer between raw screen perception and executable actions, opening a practical pathway toward reusable mobile task automation.

\textbf{Limitations.} Our framework still depends on the quality of operation-related keyframe extraction: when demonstrations contain long loading animations, subtle visual changes, or ambiguous intermediate states, the extracted keyframes may not perfectly correspond to actual human operations. In addition, while Teach VLM largely resolves high-level procedural planning, final task success on hard dynamic tasks still relies on complementary low-level capabilities of the executor, including precise element grounding, robust cross-screen memory, accurate temporal filtering, and numerical answer verification, echoing the failure patterns observed in Section~\ref{sec:case_study}. Finally, our benchmark and experiments focus on Android applications; extending the framework to broader digital environments such as desktop software, web applications, and cross-device workflows, as well as incorporating more systematic human-preference feedback and multilingual demonstrations, remains an important direction for future work.

\bibliography{main}

@article{wu2024atlas, 
  title={OS-ATLAS: A Foundation Action Model for Generalist GUI Agents}, 
  author={Wu, Zhiyong and Wu, Zhenyu and Xu, Fangzhi and Wang, Yian and Sun, Qiushi and Jia, Chengyou and Cheng, Kanzhi and Ding, Zichen and Chen, Liheng and Liang, Paul Pu and others}, 
  journal={CoRR}, 
  year={2024}
}

@misc{cheng2024seeclick,
      title={SeeClick: Harnessing GUI Grounding for Advanced Visual GUI Agents}, 
      author={Kanzhi Cheng and Qiushi Sun and Yougang Chu and Fangzhi Xu and Yantao Li and Jianbing Zhang and Zhiyong Wu},
      year={2024},
      eprint={2401.10935},
      archivePrefix={arXiv},
      primaryClass={cs.HC}
}

@inproceedings{liu2023visual,
  title={Visual instruction tuning},
  author={Liu, Haotian and Li, Chunyuan and Wu, Qingyang and Lee, Yong Jae},
  booktitle={Advances in Neural Information Processing Systems},
  volume={36},
  pages={34892--34916},
  year={2023}
}

@article{reid2024gemini,
  title={Gemini 1.5: Unlocking multimodal understanding across millions of tokens of context},
  author={Reid, Machel and Savinov, Nikolay and Denisov, Denis and Fiedel, Noah and others},
  journal={arXiv preprint arXiv:2403.05530},
  year={2024}
}

@misc{deepseekai2025deepseekv3technicalreport,
      title={DeepSeek-V3 Technical Report}, 
      author={DeepSeek-AI and Aixin Liu and Bei Feng and Bing Xue and Bingxuan Wang and Bochao Wu and Chengda Lu and Chenggang Zhao and Chengqi Deng and Chenyu Zhang and Chong Ruan and Damai Dai and Daya Guo and Dejian Yang and Deli Chen and Dongjie Ji and Erhang Li and Fangyun Lin and Fucong Dai and Fuli Luo and Guangbo Hao and Guanting Chen and Guowei Li and H. Zhang and Han Bao and Hanwei Xu and Haocheng Wang and Haowei Zhang and Honghui Ding and Huajian Xin and Huazuo Gao and Hui Li and Hui Qu and J. L. Cai and Jian Liang and Jianzhong Guo and Jiaqi Ni and Jiashi Li and Jiawei Wang and Jin Chen and Jingchang Chen and Jingyang Yuan and Junjie Qiu and Junlong Li and Junxiao Song and Kai Dong and Kai Hu and Kaige Gao and Kang Guan and Kexin Huang and Kuai Yu and Lean Wang and Lecong Zhang and Lei Xu and Leyi Xia and Liang Zhao and Litong Wang and Liyue Zhang and Meng Li and Miaojun Wang and Mingchuan Zhang and Minghua Zhang and Minghui Tang and Mingming Li and Ning Tian and Panpan Huang and Peiyi Wang and Peng Zhang and Qiancheng Wang and Qihao Zhu and Qinyu Chen and Qiushi Du and R. J. Chen and R. L. Jin and Ruiqi Ge and Ruisong Zhang and Ruizhe Pan and Runji Wang and Runxin Xu and Ruoyu Zhang and Ruyi Chen and S. S. Li and Shanghao Lu and Shangyan Zhou and Shanhuang Chen and Shaoqing Wu and Shengfeng Ye and Shengfeng Ye and Shirong Ma and Shiyu Wang and Shuang Zhou and Shuiping Yu and Shunfeng Zhou and Shuting Pan and T. Wang and Tao Yun and Tian Pei and Tianyu Sun and W. L. Xiao and Wangding Zeng and Wanjia Zhao and Wei An and Wen Liu and Wenfeng Liang and Wenjun Gao and Wenqin Yu and Wentao Zhang and X. Q. Li and Xiangyue Jin and Xianzu Wang and Xiao Bi and Xiaodong Liu and Xiaohan Wang and Xiaojin Shen and Xiaokang Chen and Xiaokang Zhang and Xiaosha Chen and Xiaotao Nie and Xiaowen Sun and Xiaoxiang Wang and Xin Cheng and Xin Liu and Xin Xie and Xingchao Liu and Xingkai Yu and Xinnan Song and Xinxia Shan and Xinyi Zhou and Xinyu Yang and Xinyuan Li and Xuecheng Su and Xuheng Lin and Y. K. Li and Y. Q. Wang and Y. X. Wei and Y. X. Zhu and Yang Zhang and Yanhong Xu and Yanhong Xu and Yanping Huang and Yao Li and Yao Zhao and Yaofeng Sun and Yaohui Li and Yaohui Wang and Yi Yu and Yi Zheng and Yichao Zhang and Yifan Shi and Yiliang Xiong and Ying He and Ying Tang and Yishi Piao and Yisong Wang and Yixuan Tan and Yiyang Ma and Yiyuan Liu and Yongqiang Guo and Yu Wu and Yuan Ou and Yuchen Zhu and Yuduan Wang and Yue Gong and Yuheng Zou and Yujia He and Yukun Zha and Yunfan Xiong and Yunxian Ma and Yuting Yan and Yuxiang Luo and Yuxiang You and Yuxuan Liu and Yuyang Zhou and Z. F. Wu and Z. Z. Ren and Zehui Ren and Zhangli Sha and Zhe Fu and Zhean Xu and Zhen Huang and Zhen Zhang and Zhenda Xie and Zhengyan Zhang and Zhewen Hao and Zhibin Gou and Zhicheng Ma and Zhigang Yan and Zhihong Shao and Zhipeng Xu and Zhiyu Wu and Zhongyu Zhang and Zhuoshu Li and Zihui Gu and Zijia Zhu and Zijun Liu and Zilin Li and Ziwei Xie and Ziyang Song and Ziyi Gao and Zizheng Pan},
      year={2025},
      eprint={2412.19437},
      archivePrefix={arXiv},
      primaryClass={cs.CL},
      url={https://arxiv.org/abs/2412.19437}, 
}

@misc{wang2025internvl35advancingopensourcemultimodal,
      title={InternVL3.5: Advancing Open-Source Multimodal Models in Versatility, Reasoning, and Efficiency}, 
      author={Weiyun Wang and Zhangwei Gao and Lixin Gu and Hengjun Pu and Long Cui and Xingguang Wei and Zhaoyang Liu and Linglin Jing and Shenglong Ye and Jie Shao and Zhaokai Wang and Zhe Chen and Hongjie Zhang and Ganlin Yang and Haomin Wang and Qi Wei and Jinhui Yin and Wenhao Li and Erfei Cui and Guanzhou Chen and Zichen Ding and Changyao Tian and Zhenyu Wu and Jingjing Xie and Zehao Li and Bowen Yang and Yuchen Duan and Xuehui Wang and Zhi Hou and Haoran Hao and Tianyi Zhang and Songze Li and Xiangyu Zhao and Haodong Duan and Nianchen Deng and Bin Fu and Yinan He and Yi Wang and Conghui He and Botian Shi and Junjun He and Yingtong Xiong and Han Lv and Lijun Wu and Wenqi Shao and Kaipeng Zhang and Huipeng Deng and Biqing Qi and Jiaye Ge and Qipeng Guo and Wenwei Zhang and Songyang Zhang and Maosong Cao and Junyao Lin and Kexian Tang and Jianfei Gao and Haian Huang and Yuzhe Gu and Chengqi Lyu and Huanze Tang and Rui Wang and Haijun Lv and Wanli Ouyang and Limin Wang and Min Dou and Xizhou Zhu and Tong Lu and Dahua Lin and Jifeng Dai and Weijie Su and Bowen Zhou and Kai Chen and Yu Qiao and Wenhai Wang and Gen Luo},
      year={2025},
      eprint={2508.18265},
      archivePrefix={arXiv},
      primaryClass={cs.CV},
      url={https://arxiv.org/abs/2508.18265}, 
}

@misc{yang2023autogptonlinedecisionmaking,
      title={Auto-GPT for Online Decision Making: Benchmarks and Additional Opinions}, 
      author={Hui Yang and Sifu Yue and Yunzhong He},
      year={2023},
      eprint={2306.02224},
      archivePrefix={arXiv},
      primaryClass={cs.AI},
      url={https://arxiv.org/abs/2306.02224}, 
}

@misc{qin2025uitarspioneeringautomatedgui,
      title={UI-TARS: Pioneering Automated GUI Interaction with Native Agents}, 
      author={Yujia Qin and Yining Ye and Junjie Fang and Haoming Wang and Shihao Liang and Shizuo Tian and Junda Zhang and Jiahao Li and Yunxin Li and Shijue Huang and Wanjun Zhong and Kuanye Li and Jiale Yang and Yu Miao and Woyu Lin and Longxiang Liu and Xu Jiang and Qianli Ma and Jingyu Li and Xiaojun Xiao and Kai Cai and Chuang Li and Yaowei Zheng and Chaolin Jin and Chen Li and Xiao Zhou and Minchao Wang and Haoli Chen and Zhaojian Li and Haihua Yang and Haifeng Liu and Feng Lin and Tao Peng and Xin Liu and Guang Shi},
      year={2025},
      eprint={2501.12326},
      archivePrefix={arXiv},
      primaryClass={cs.AI},
      url={https://arxiv.org/abs/2501.12326}, 
}

@misc{ye2025mobileagentv3fundamentalagentsgui,
      title={Mobile-Agent-v3: Fundamental Agents for GUI Automation}, 
      author={Jiabo Ye and Xi Zhang and Haiyang Xu and Haowei Liu and Junyang Wang and Zhaoqing Zhu and Ziwei Zheng and Feiyu Gao and Junjie Cao and Zhengxi Lu and Jitong Liao and Qi Zheng and Fei Huang and Jingren Zhou and Ming Yan},
      year={2025},
      eprint={2508.15144},
      archivePrefix={arXiv},
      primaryClass={cs.AI},
      url={https://arxiv.org/abs/2508.15144}, 
}

@misc{hong2024metagptmetaprogrammingmultiagent,
      title={MetaGPT: Meta Programming for A Multi-Agent Collaborative Framework}, 
      author={Sirui Hong and Mingchen Zhuge and Jiaqi Chen and Xiawu Zheng and Yuheng Cheng and Ceyao Zhang and Jinlin Wang and Zili Wang and Steven Ka Shing Yau and Zijuan Lin and Liyang Zhou and Chenyu Ran and Lingfeng Xiao and Chenglin Wu and Jürgen Schmidhuber},
      year={2024},
      eprint={2308.00352},
      archivePrefix={arXiv},
      primaryClass={cs.AI},
      url={https://arxiv.org/abs/2308.00352}, 
}

@misc{zhang2023appagentmultimodalagentssmartphone,
      title={AppAgent: Multimodal Agents as Smartphone Users}, 
      author={Chi Zhang and Zhao Yang and Jiaxuan Liu and Yucheng Han and Xin Chen and Zebiao Huang and Bin Fu and Gang Yu},
      year={2023},
      eprint={2312.13771},
      archivePrefix={arXiv},
      primaryClass={cs.CV},
      url={https://arxiv.org/abs/2312.13771}, 
}

@misc{wang2025opencuaopenfoundationscomputeruse,
      title={OpenCUA: Open Foundations for Computer-Use Agents}, 
      author={Xinyuan Wang and Bowen Wang and Dunjie Lu and Junlin Yang and Tianbao Xie and Junli Wang and Jiaqi Deng and Xiaole Guo and Yiheng Xu and Chen Henry Wu and Zhennan Shen and Zhuokai Li and Ryan Li and Xiaochuan Li and Junda Chen and Boyuan Zheng and Peihang Li and Fangyu Lei and Ruisheng Cao and Yeqiao Fu and Dongchan Shin and Martin Shin and Jiarui Hu and Yuyan Wang and Jixuan Chen and Yuxiao Ye and Danyang Zhang and Dikang Du and Hao Hu and Huarong Chen and Zaida Zhou and Haotian Yao and Ziwei Chen and Qizheng Gu and Yipu Wang and Heng Wang and Diyi Yang and Victor Zhong and Flood Sung and Y. Charles and Zhilin Yang and Tao Yu},
      year={2025},
      eprint={2508.09123},
      archivePrefix={arXiv},
      primaryClass={cs.AI},
      url={https://arxiv.org/abs/2508.09123}, 
}

@misc{masterman2024landscapeemergingaiagent,
      title={The Landscape of Emerging AI Agent Architectures for Reasoning, Planning, and Tool Calling: A Survey}, 
      author={Tula Masterman and Sandi Besen and Mason Sawtell and Alex Chao},
      year={2024},
      eprint={2404.11584},
      archivePrefix={arXiv},
      primaryClass={cs.AI},
      url={https://arxiv.org/abs/2404.11584}, 
}

@misc{yao2023reactsynergizingreasoningacting,
      title={ReAct: Synergizing Reasoning and Acting in Language Models}, 
      author={Shunyu Yao and Jeffrey Zhao and Dian Yu and Nan Du and Izhak Shafran and Karthik Narasimhan and Yuan Cao},
      year={2023},
      eprint={2210.03629},
      archivePrefix={arXiv},
      primaryClass={cs.CL},
      url={https://arxiv.org/abs/2210.03629}, 
}

@misc{schick2023toolformerlanguagemodelsteach,
      title={Toolformer: Language Models Can Teach Themselves to Use Tools}, 
      author={Timo Schick and Jane Dwivedi-Yu and Roberto Dessì and Roberta Raileanu and Maria Lomeli and Luke Zettlemoyer and Nicola Cancedda and Thomas Scialom},
      year={2023},
      eprint={2302.04761},
      archivePrefix={arXiv},
      primaryClass={cs.CL},
      url={https://arxiv.org/abs/2302.04761}, 
}

@misc{openai2024gpt4technicalreport,
      title={GPT-4 Technical Report}, 
      author={OpenAI and Josh Achiam and Steven Adler and Sandhini Agarwal and Lama Ahmad and Ilge Akkaya and Florencia Leoni Aleman and Diogo Almeida and Janko Altenschmidt and Sam Altman and Shyamal Anadkat and Red Avila and Igor Babuschkin and Suchir Balaji and Valerie Balcom and Paul Baltescu and Haiming Bao and Mohammad Bavarian and Jeff Belgum and Irwan Bello and Jake Berdine and Gabriel Bernadett-Shapiro and Christopher Berner and Lenny Bogdonoff and Oleg Boiko and Madelaine Boyd and Anna-Luisa Brakman and Greg Brockman and Tim Brooks and Miles Brundage and Kevin Button and Trevor Cai and Rosie Campbell and Andrew Cann and Brittany Carey and Chelsea Carlson and Rory Carmichael and Brooke Chan and Che Chang and Fotis Chantzis and Derek Chen and Sully Chen and Ruby Chen and Jason Chen and Mark Chen and Ben Chess and Chester Cho and Casey Chu and Hyung Won Chung and Dave Cummings and Jeremiah Currier and Yunxing Dai and Cory Decareaux and Thomas Degry and Noah Deutsch and Damien Deville and Arka Dhar and David Dohan and Steve Dowling and Sheila Dunning and Adrien Ecoffet and Atty Eleti and Tyna Eloundou and David Farhi and Liam Fedus and Niko Felix and Simón Posada Fishman and Juston Forte and Isabella Fulford and Leo Gao and Elie Georges and Christian Gibson and Vik Goel and Tarun Gogineni and Gabriel Goh and Rapha Gontijo-Lopes and Jonathan Gordon and Morgan Grafstein and Scott Gray and Ryan Greene and Joshua Gross and Shixiang Shane Gu and Yufei Guo and Chris Hallacy and Jesse Han and Jeff Harris and Yuchen He and Mike Heaton and Johannes Heidecke and Chris Hesse and Alan Hickey and Wade Hickey and Peter Hoeschele and Brandon Houghton and Kenny Hsu and Shengli Hu and Xin Hu and Joost Huizinga and Shantanu Jain and Shawn Jain and Joanne Jang and Angela Jiang and Roger Jiang and Haozhun Jin and Denny Jin and Shino Jomoto and Billie Jonn and Heewoo Jun and Tomer Kaftan and Łukasz Kaiser and Ali Kamali and Ingmar Kanitscheider and Nitish Shirish Keskar and Tabarak Khan and Logan Kilpatrick and Jong Wook Kim and Christina Kim and Yongjik Kim and Jan Hendrik Kirchner and Jamie Kiros and Matt Knight and Daniel Kokotajlo and Łukasz Kondraciuk and Andrew Kondrich and Aris Konstantinidis and Kyle Kosic and Gretchen Krueger and Vishal Kuo and Michael Lampe and Ikai Lan and Teddy Lee and Jan Leike and Jade Leung and Daniel Levy and Chak Ming Li and Rachel Lim and Molly Lin and Stephanie Lin and Mateusz Litwin and Theresa Lopez and Ryan Lowe and Patricia Lue and Anna Makanju and Kim Malfacini and Sam Manning and Todor Markov and Yaniv Markovski and Bianca Martin and Katie Mayer and Andrew Mayne and Bob McGrew and Scott Mayer McKinney and Christine McLeavey and Paul McMillan and Jake McNeil and David Medina and Aalok Mehta and Jacob Menick and Luke Metz and Andrey Mishchenko and Pamela Mishkin and Vinnie Monaco and Evan Morikawa and Daniel Mossing and Tong Mu and Mira Murati and Oleg Murk and David Mély and Ashvin Nair and Reiichiro Nakano and Rajeev Nayak and Arvind Neelakantan and Richard Ngo and Hyeonwoo Noh and Long Ouyang and Cullen O'Keefe and Jakub Pachocki and Alex Paino and Joe Palermo and Ashley Pantuliano and Giambattista Parascandolo and Joel Parish and Emy Parparita and Alex Passos and Mikhail Pavlov and Andrew Peng and Adam Perelman and Filipe de Avila Belbute Peres and Michael Petrov and Henrique Ponde de Oliveira Pinto and Michael and Pokorny and Michelle Pokrass and Vitchyr H. Pong and Tolly Powell and Alethea Power and Boris Power and Elizabeth Proehl and Raul Puri and Alec Radford and Jack Rae and Aditya Ramesh and Cameron Raymond and Francis Real and Kendra Rimbach and Carl Ross and Bob Rotsted and Henri Roussez and Nick Ryder and Mario Saltarelli and Ted Sanders and Shibani Santurkar and Girish Sastry and Heather Schmidt and David Schnurr and John Schulman and Daniel Selsam and Kyla Sheppard and Toki Sherbakov and Jessica Shieh and Sarah Shoker and Pranav Shyam and Szymon Sidor and Eric Sigler and Maddie Simens and Jordan Sitkin and Katarina Slama and Ian Sohl and Benjamin Sokolowsky and Yang Song and Natalie Staudacher and Felipe Petroski Such and Natalie Summers and Ilya Sutskever and Jie Tang and Nikolas Tezak and Madeleine B. Thompson and Phil Tillet and Amin Tootoonchian and Elizabeth Tseng and Preston Tuggle and Nick Turley and Jerry Tworek and Juan Felipe Cerón Uribe and Andrea Vallone and Arun Vijayvergiya and Chelsea Voss and Carroll Wainwright and Justin Jay Wang and Alvin Wang and Ben Wang and Jonathan Ward and Jason Wei and CJ Weinmann and Akila Welihinda and Peter Welinder and Jiayi Weng and Lilian Weng and Matt Wiethoff and Dave Willner and Clemens Winter and Samuel Wolrich and Hannah Wong and Lauren Workman and Sherwin Wu and Jeff Wu and Michael Wu and Kai Xiao and Tao Xu and Sarah Yoo and Kevin Yu and Qiming Yuan and Wojciech Zaremba and Rowan Zellers and Chong Zhang and Marvin Zhang and Shengjia Zhao and Tianhao Zheng and Juntang Zhuang and William Zhuk and Barret Zoph},
      year={2024},
      eprint={2303.08774},
      archivePrefix={arXiv},
      primaryClass={cs.CL},
      url={https://arxiv.org/abs/2303.08774}, 
}

@article{10.1093/nsr/nwae403,
    author = {Yin, Shukang and Fu, Chaoyou and Zhao, Sirui and Li, Ke and Sun, Xing and Xu, Tong and Chen, Enhong},
    title = {A survey on multimodal large language models},
    journal = {National Science Review},
    volume = {11},
    number = {12},
    pages = {nwae403},
    year = {2024},
    month = {12},
    issn = {2095-5138},
    doi = {10.1093/nsr/nwae403},
    url = {https://doi.org/10.1093/nsr/nwae403},
    eprint = {https://academic.oup.com/nsr/article-pdf/11/12/nwae403/61201557/nwae403.pdf},
}

@inproceedings{ZhangWTLXXWT24,
  author       = {Jiwen Zhang and
                  Jihao Wu and
                  Yihua Teng and
                  Minghui Liao and
                  Nuo Xu and
                  Xiao Xiao and
                  Zhongyu Wei and
                  Duyu Tang},
  title        = {Android in the Zoo: Chain-of-Action-Thought for {GUI} Agents},
  booktitle    = {Findings of the Association for Computational Linguistics: {EMNLP}
                  2024, Miami, Florida, USA, November 12-16, 2024},
  pages        = {12016--12031},
  year         = {2024},
  url          = {https://doi.org/10.18653/v1/2024.findings-emnlp.702},
  doi          = {10.18653/V1/2024.FINDINGS-EMNLP.702},
  bibsource    = {dblp computer science bibliography, https://dblp.org}
}

@inproceedings{lu2025guiodyssey,
  title={GUIOdyssey: A comprehensive dataset for cross-app GUI navigation on mobile devices},
  author={Lu, Quanfeng and Shao, Wenqi and Liu, Zitao and Du, Lingxiao and Meng, Fanqing and Li, Boxuan and Chen, Botong and Huang, Siyuan and Zhang, Kaipeng and Luo, Ping},
  booktitle={Proceedings of the IEEE/CVF International Conference on Computer Vision},
  pages={22404--22414},
  year={2025}
}

@misc{rawles2024androidworlddynamicbenchmarkingenvironment,
      title={AndroidWorld: A Dynamic Benchmarking Environment for Autonomous Agents},
      author={Christopher Rawles and Sarah Clinckemaillie and Yifan Chang and Jonathan Waltz and Gabrielle Lau and Marybeth Fair and Alice Li and William Bishop and Wei Li and Folawiyo Campbell-Ajala and Daniel Toyama and Robert Berry and Divya Tyamagundlu and Timothy Lillicrap and Oriana Riva},
      year={2024},
      eprint={2405.14573},
      archivePrefix={arXiv},
      primaryClass={cs.AI},
      url={https://arxiv.org/abs/2405.14573},
}

@misc{qwen3vltechnicalreport,
      title={Qwen3-VL Technical Report},
      author={Bai, Shuai and Cai, Yuxuan and Chen, Ruizhe and others},
      year={2025},
      eprint={2511.21631},
      archivePrefix={arXiv},
      primaryClass={cs.CV},
      doi={10.48550/arXiv.2511.21631},
      url={https://arxiv.org/abs/2511.21631},
}

@misc{qwen3vl8binstruct,
      title={Qwen3-VL-8B-Instruct},
      author={{Qwen Team}},
      year={2025},
      howpublished={\url{https://huggingface.co/Qwen/Qwen3-VL-8B-Instruct}},
      note={Model card, accessed 2026-05-18},
}

@misc{zhao2024swiftascalablelightweightinfrastructure,
      title={SWIFT:A Scalable lightWeight Infrastructure for Fine-Tuning},
      author={Yuze Zhao and Jintao Huang and Jinghan Hu and Xingjun Wang and Yunlin Mao and Daoze Zhang and Zeyinzi Jiang and Zhikai Wu and Baole Ai and Ang Wang and Wenmeng Zhou and Yingda Chen},
      year={2024},
      eprint={2408.05517},
      archivePrefix={arXiv},
      primaryClass={cs.CL},
      url={https://arxiv.org/abs/2408.05517},
}

@misc{xia2026skillrl,
      title={SkillRL: Evolving Agents via Recursive Skill-Augmented Reinforcement Learning},
      author={Peng Xia and Jianwen Chen and Hanyang Wang and Jiaqi Liu and Kaide Zeng and Yu Wang and Siwei Han and Yiyang Zhou and Xujiang Zhao and Haifeng Chen and Zeyu Zheng and Cihang Xie and Huaxiu Yao},
      year={2026},
      eprint={2602.08234},
      archivePrefix={arXiv},
      primaryClass={cs.LG},
      url={https://arxiv.org/abs/2602.08234},
}

@misc{lu2026skill0,
      title={SKILL0: In-Context Agentic Reinforcement Learning for Skill Internalization},
      author={Zhengxi Lu and Zhiyuan Yao and Jinyang Wu and Chengcheng Han and Qi Gu and Xunliang Cai and Weiming Lu and Jun Xiao and Yueting Zhuang and Yongliang Shen},
      year={2026},
      eprint={2604.02268},
      archivePrefix={arXiv},
      primaryClass={cs.LG},
      url={https://arxiv.org/abs/2604.02268},
}

@misc{wang2026skillsd,
      title={Skill-SD: Skill-Conditioned Self-Distillation for Multi-turn LLM Agents},
      author={Hao Wang and Guozhi Wang and Han Xiao and Yufeng Zhou and Yue Pan and Jichao Wang and Ke Xu and Yafei Wen and Xiaohu Ruan and Xiaoxin Chen and Honggang Qi},
      year={2026},
      eprint={2604.10674},
      archivePrefix={arXiv},
      primaryClass={cs.AI},
      url={https://arxiv.org/abs/2604.10674},
}

@misc{openai2024gpt4osystemcard,
      title={GPT-4o System Card},
      author={{OpenAI}},
      year={2024},
      eprint={2410.21276},
      archivePrefix={arXiv},
      primaryClass={cs.CL},
      url={https://arxiv.org/abs/2410.21276},
}

@misc{bytedanceseed2026seed18,
      title={Seed1.8 Model Card: Towards Generalized Real-World Agency},
      author={{ByteDance Seed}},
      year={2026},
      eprint={2603.20633},
      archivePrefix={arXiv},
      primaryClass={cs.AI},
      url={https://arxiv.org/abs/2603.20633},
}

@misc{bytedanceseed2026seed20,
      title={Seed2.0 Model Card},
      author={{ByteDance Seed}},
      year={2026},
      howpublished={\url{https://seed.bytedance.com/en/seed2}},
      note={Official model card and model page, accessed 2026-06-05},
}

@misc{qwen2026qwen35,
      title={Qwen 3.5 Technical Report},
      author={{Qwen Team}},
      year={2026},
      howpublished={Technical report},
}
\clearpage
\section{Appendix}~\label{sec:appendix}

\subsection{Teach VLM Training Prompt}
\label{sec:appendix_training_prompt}

Teach VLM is trained to translate visual state transitions into natural-language
operation descriptions. In the pure-visual setting used for the final training
round, the input contains screenshots only and does not contain structured
\texttt{tool\_call} annotations. A single-step training example is constructed
from two consecutive screenshots:

\begin{quote}
\small
\begin{verbatim}
You are given two consecutive screenshots.
Image 1 is before the operation.
Image 2 is after the operation.
Task instruction: <task instruction>
Infer the operation between the screenshots.
Describe only the operation itself.
Do not describe the screen change result.
Focus on action, target element, and text.
Mention scroll direction when applicable.
Do not mention coordinates.
Output concise English only.
Return exactly one <conclusion> tag.
<image>
<image>
\end{verbatim}
\end{quote}

The target response uses a single conclusion tag:

\begin{quote}
\small
\begin{verbatim}
<conclusion>
click on the search bar.
</conclusion>
\end{verbatim}
\end{quote}

For sequence-level supervision, Teach VLM receives a chronological sequence of
screenshots or multiple before/after screenshot pairs and is asked to output one
\texttt{<conclusion>} block for each operation in order:

\begin{quote}
\small
\begin{verbatim}
You are given before/after screenshot pairs.
There are N operations and 2N screenshots.
Pair 1: Image 1 is before, Image 2 is after.
...
Infer the operation for each pair.
Each step is one concise English action.
Focus on operations, not screen effects.
Do not mention coordinates.
Return one <conclusion> tag per operation.
<image>
...
<image>
\end{verbatim}
\end{quote}

\subsection{Example of Operational Knowledge}
\label{sec:appendix_operational_knowledge}

The following example shows an operation sequence generated by Teach VLM from a
mobile screen trajectory. It is presented as step-wise operational knowledge,
where each sentence describes the intended operation rather than the visual
effect after the operation.

\begin{quote}
\small
\begin{verbatim}
Task: Google the capital of Canada

Operational knowledge:
1. scroll up
2. click on the Google app
3. click on the search bar
4. type "capital of Canada"
5. click on the search result
6. stop and set the query as completed
\end{verbatim}
\end{quote}

This example illustrates the form of procedural guidance used in
Teach-and-Repeat: the knowledge is concise, ordered, and expressed in natural
language, making it directly readable by both humans and downstream
screen-based agents.

\subsection{Example of Injecting Operational Knowledge into Android World}
\label{sec:appendix_knowledge_injection}

In Android World, operational knowledge is injected as an auxiliary reference
strategy in the agent prompt. The injected text is not treated as an executable
policy. Instead, it provides high-level procedural guidance while the agent
still grounds each decision in the current screenshots and accessibility tree.

\begin{quote}
\small
\begin{verbatim}
Reference strategy:
Reference Goal: Google the capital of Canada
Reference Action Descriptions:
1. scroll up
2. click on the Google app
3. click on the search bar
4. type "capital of Canada"
5. click on the search result
6. stop and set the query as completed
Use these descriptions as high-level guidance.
Do not copy the sequence blindly.
Ground decisions in current screenshots.
\end{verbatim}
\end{quote}

This prompt block is inserted before the agent makes the next action decision.
The design separates perception-derived operation knowledge from closed-loop
execution: Teach VLM supplies a reusable strategy, while the Android World agent
remains responsible for selecting the concrete action under the current screen
state.


\end{document}